\documentclass[review, compsoc]{IEEEtran}
\usepackage{amsfonts}
\usepackage{amsmath}
\usepackage{amsmath,epsfig,amsmath}
\usepackage{amsfonts,amssymb,amsthm}
\usepackage{epsfig}
\usepackage{float}
\usepackage{graphicx}
\usepackage{color,xcolor,hyperref}
\usepackage{multirow}
\usepackage{tipa}
\usepackage{picins}
\usepackage{color}
\usepackage{booktabs}
\usepackage{threeparttable}
\usepackage{makecell}

\begin{document}

\title{Vision-Based Traffic Accident \\Detection and Anticipation: A Survey}

\author{Jianwu Fang, Jiahuan Qiao, Jianru Xue, and Zhengguo Li

%\author{Jianwu Fang, Jiahuan Qiao, Jianru Xue, Zhengguo Li, and Tat-Seng Chua

%\author{Jianwu Fang
\thanks{
J. Fang is with the College of Transportation Engineering, Chang'an University, Xi'an, China, and a visiting scholar at the NExT++ Research Centre of the School of Computing, National University of Singapore, Singapore.
        {(fangjianwu@ieee.org)}.} 
         \thanks{J. Qiao is with the Applied Mathematics and Artificial Intelligence, Institute of Information Technologies and Computer Science, National Research University Moscow Power Engineering Institute, Russia (TsiaoTs@mpei.ru).}
% \thanks{T-S. Chua is with the Sea-NExT Joint Research Centre of the School of Computing, National University of Singapore, Singapore.
%    {(\{dcscts\}@nus.edu.sg)}.}
\thanks{J. Xue is with the Institute of Artificial Intelligence and Robotics, Xi'an Jiaotong University, Xi'an, China
     {(jrxue@mail.xjtu.edu.cn).}}
 \thanks{Z. Li is with the Institute for Infocomm Research, Agency for Science, Technology and Research (A$^*$STAR), Singapore
 {(ezgli@i2r.a-star.edu.sg).}}
}

% The paper headers
\markboth{IEEE Latex}%
{Shell \MakeLowercase{\textit{et al.}}: Bare Demo of IEEEtran.cls for Computer Society Journals}
% The only time the second header will appear is for the odd numbered pages
% after the title page when using the twoside option.
% 
% *** Note that you probably will NOT want to include the author's ***
% *** name in the headers of peer review papers.                   ***
% You can use \ifCLASSOPTIONpeerreview for conditional compilation here if
% you desire.

\IEEEtitleabstractindextext{%
\begin{abstract}
Traffic accident detection and anticipation is an obstinate road safety problem and painstaking efforts have been devoted. With the rapid growth of video data, Vision-based Traffic Accident Detection and Anticipation (named Vision-TAD and Vision-TAA) become the last one-mile problem for safe driving and surveillance safety. However, the long-tailed, unbalanced, highly dynamic, complex, and uncertain properties of traffic accidents form the Out-of-Distribution (OOD) feature for Vision-TAD and Vision-TAA. Current AI development may focus on these OOD but important problems. What has been done for Vision-TAD and Vision-TAA? What direction we should focus on in the future for this problem? A comprehensive survey is important. We present the first survey on Vision-TAD in the deep learning era and the first-ever survey for Vision-TAA. The pros and cons of each research prototype are discussed in detail during the investigation. In addition, we also provide a critical review of 31 publicly available benchmarks and related evaluation metrics. Through this survey, we want to spawn new insights and open possible trends for Vision-TAD and Vision-TAA tasks.
\end{abstract}

% Note that keywords are not normally used for peerreview papers.
\begin{IEEEkeywords}
Traffic accident detection and anticipation, surveillance safety, safe driving, autoencoder, benchmarks
%\vspace{4em}
\end{IEEEkeywords}}

% make the title area
\maketitle

% To allow for easy dual compilation without having to reenter the
% abstract/keywords data, the \IEEEtitleabstractindextext text will
% not be used in maketitle, but will appear (i.e., to be "transported")
% here as \IEEEdisplaynontitleabstractindextext when the compsoc 
% or transmag modes are not selected <OR> if conference mode is selected 
% - because all conference papers position the abstract like regular
% papers do.
\IEEEdisplaynontitleabstractindextext
% \IEEEdisplaynontitleabstractindextext has no effect when using
% compsoc or transmag under a non-conference mode.

% For peer review papers, you can put extra information on the cover
% page as needed:
% \ifCLASSOPTIONpeerreview
% \begin{center} \bfseries EDICS Category: 3-BBND \end{center}
% \fi
%
% For peerreview papers, this IEEEtran command inserts a page break and
% creates the second title. It will be ignored for other modes.
\IEEEpeerreviewmaketitle

\IEEEraisesectionheading{
\section{Introduction}
\label{section1}}
\IEEEPARstart{V}{ast} amount of road accidents have deprived many human lives each year. More than half of the deaths are vulnerable road users (\emph{i.e.,} pedestrians, cyclists, and motorbikes), and road accidents are the leading killer for young people aged from 5 to 29 years old \cite{trafficdeath}. This painful fact drives lots of researchers and institutes to devote their efforts to developing traffic accident detection or anticipation systems. Fortunately, ubiquitous cameras in road surveillance and dashcam views have contributed to a nearly transparent world with the recording of each traffic accident. Thus, the abundant video data provides a basis for Vision-based Traffic Accident Detection and Anticipation, named Vision-TAD and Vision-TAA, respectively. Consequently, the rapid growth of traffic video data provides unprecedented sources for searching the margin of traffic accidents in normal situations. 
 \begin{figure}[!t]
  \centering
 \includegraphics[width=\hsize]{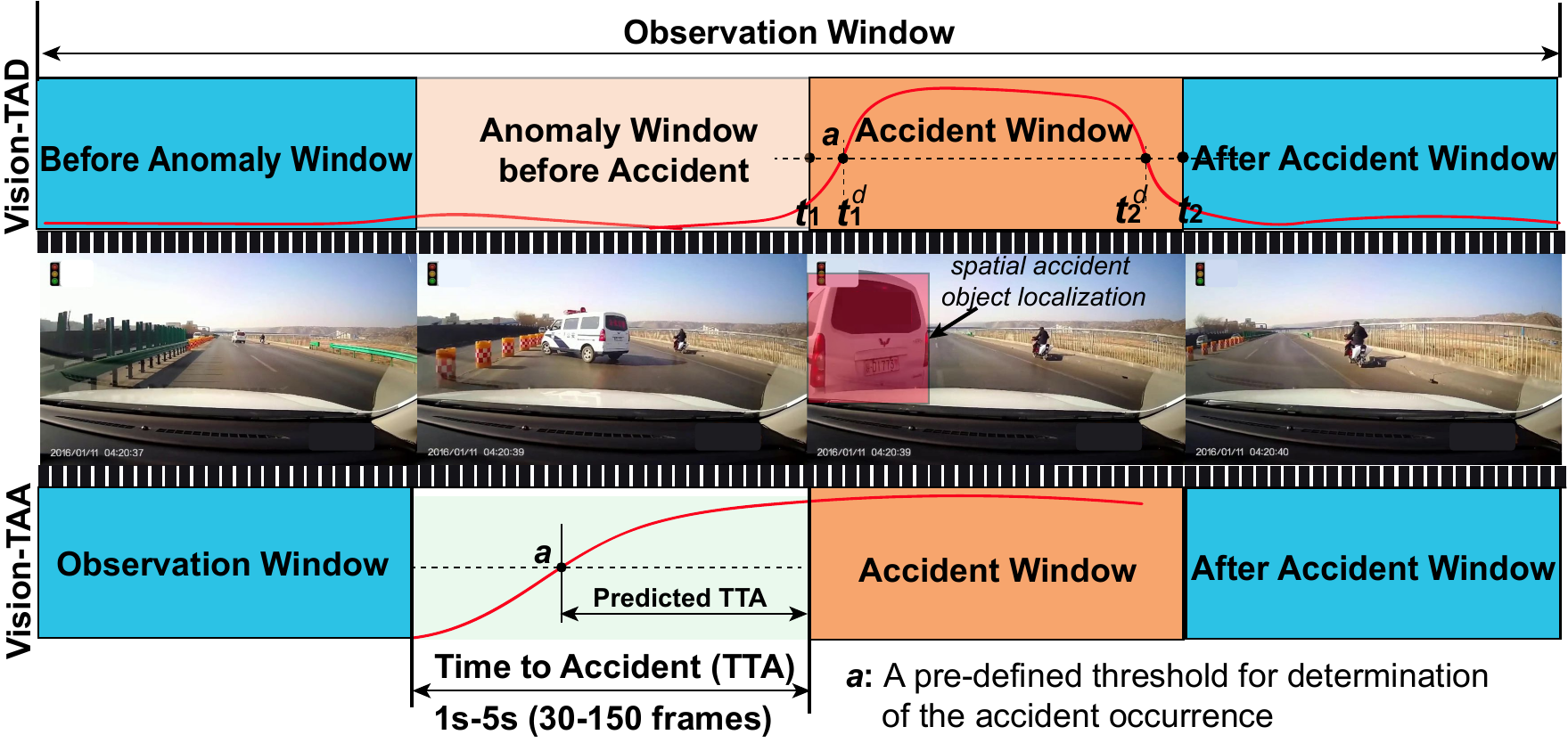}
  \caption{The illustration for Vision-TAD and Vision-TAA with a dashcam accident video, where Vision-TAD involves the spatial-temporal localization of the accident and prefers the determined accident window [$t_1^d$, $t_2^d$] to be approximate to [$t_1$, $t_2$], while Vision-TAA pursues an early prediction of the accident with a large time-to-accident (Predicted-TAA $\tau$). Notably, the observation windows of Vision-TAD and Vision-TAA are different, and Vision-TAD is easily confounded by traffic anomaly detection.}
  \label{fig1}
\end{figure}

\textbf{Traffic Accident vs. Traffic Anomaly}: Notably, there is one confusion between traffic anomaly detection and traffic accident detection. They have different temporal localization windows and the anomaly window is larger than the accident window. In addition, traffic accidents have a clearer definition than traffic anomalies, where crashes occur. On the contrary, traffic anomaly owns wider situations, in which irregular movements without the possibility of a crash are also involved, and traffic accidents are special cases of traffic anomaly. 

As shown by a dashcam video in Fig. \ref{fig1}, given an observation frame window, Vision-TAD is formulated as the localization of the time window and spatial regions where the accident occurs, while Vision-TAA prefers an early warning for the future accident with a time interval of Time-to-Accident (TTA) $\tau$ from the current time. The formulation difference makes Vision-TAD and Vision-TAA own differing temporal relation modeling in video frames. Similarly, the number of accident frames in Vision-TAD and Vision-TAA usually occupies a small ratio of each video sequence. Coupling with various interfering factors (severe light or weather conditions, small scale of the crashing objects, \emph{etc.}), the detection or anticipation of traffic accidents is not an easy task. Even so, the researchers meticulously design many models to look for useful patterns and various deep-learning models become the primary choice for this task, with the alternatives of supervised \cite{DBLP:journals/tits/KamijoMIS00,dogru2018traffic,DBLP:conf/icpr/FatimaKK20,DBLP:journals/corr/abs-1910-11072}, semi-supervised \cite{DBLP:journals/eswa/HuWCG21,chakraborty2018freeway}, weakly-supervised \cite{DBLP:journals/tip/LvZCXLY21,DBLP:conf/ijcai/0030ZLW0LDL21}, and unsupervised learning \cite{DBLP:journals/tcyb/YuanFW15,DBLP:journals/tcsv/LiFXX21} frameworks. 
Accurate Vision-TAD and Vision-TAA face various \textbf{challenges}. 

 \begin{figure}[!t]
	\centering
	\includegraphics[width=\hsize]{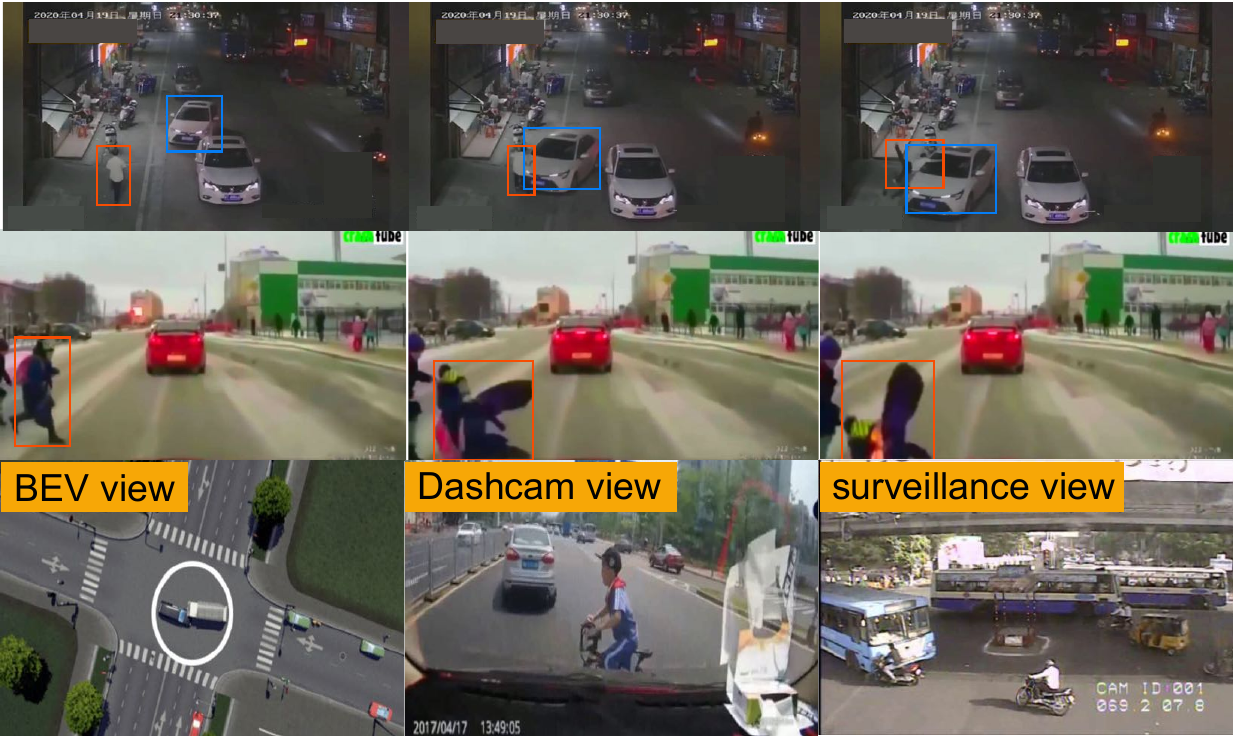}
	\caption{Some traffic accident cases in surveillance and dashcam videos, where crashing objects may undergo drastic shape change or harsh environment.}
	\label{fig2}
\end{figure}

\textbf{1) Long-tailed and Safe-critical Property.} No one road user wants to be involved in a road accident, which forms the rarity of the traffic accident from the time window and spatial locations. Usually, an accident occurs suddenly and may only contain about 20-60 video frames (1-2 seconds), where the road users (pedestrians, cyclists, cars, \emph{etc.})  take a small spatial ratio of the whole frame. Therefore, the normal background frames and appearance cause a confounding issue \cite{li2022invariant} for the causal part grounding of accidents.

\textbf{2) Complex Scene Evolution.} The objects in traffic accidents often show a drastic change in the shape, location, and relations of the participants. As shown in Fig. \ref{fig2}, a pedestrian undergoes a ``crossing" behavior, and only part of the body is visible when he is hit by a moving car. All these situations cause a complex scene evolution. With this challenge, the Vision-TAD and Vision-TAA must comprise powerful motion or appearance change encoding modules because the detection and tracking pipeline may be invalid.

\textbf{3) Harsh Environment.} Severe weather and low light conditions weaken the clearness of the road participants in the vision perception systems. Therefore, many low-level vision works of dehazing \cite{li2018end,li2021multi,li2022dual}, mixed weather removal~\cite{wang2023context}, \emph{etc.}, are proposed. Facing harsh environments, these low-level vision tasks cannot highlight the details well and the quality improvement is still far from satisfactory. 

\textbf{4) Determination Uncertainty.} Because of the rarity of the traffic accident data, Vision-TAD and Vision-TAA own the natural Algebraic Uncertainty~\cite{bao2020uncertainty,Zheng_2021,chen2022composed} (\emph{i.e.,} the determination concerns with the uncertainty of latent variables, such as observation noise and data insufficiency) and Epistemic Uncertainty~\cite{gal2016dropout} (\emph{i.e.,} model generalization). 
\begin{figure}[!t]
	\centering
	\includegraphics[width=\hsize]{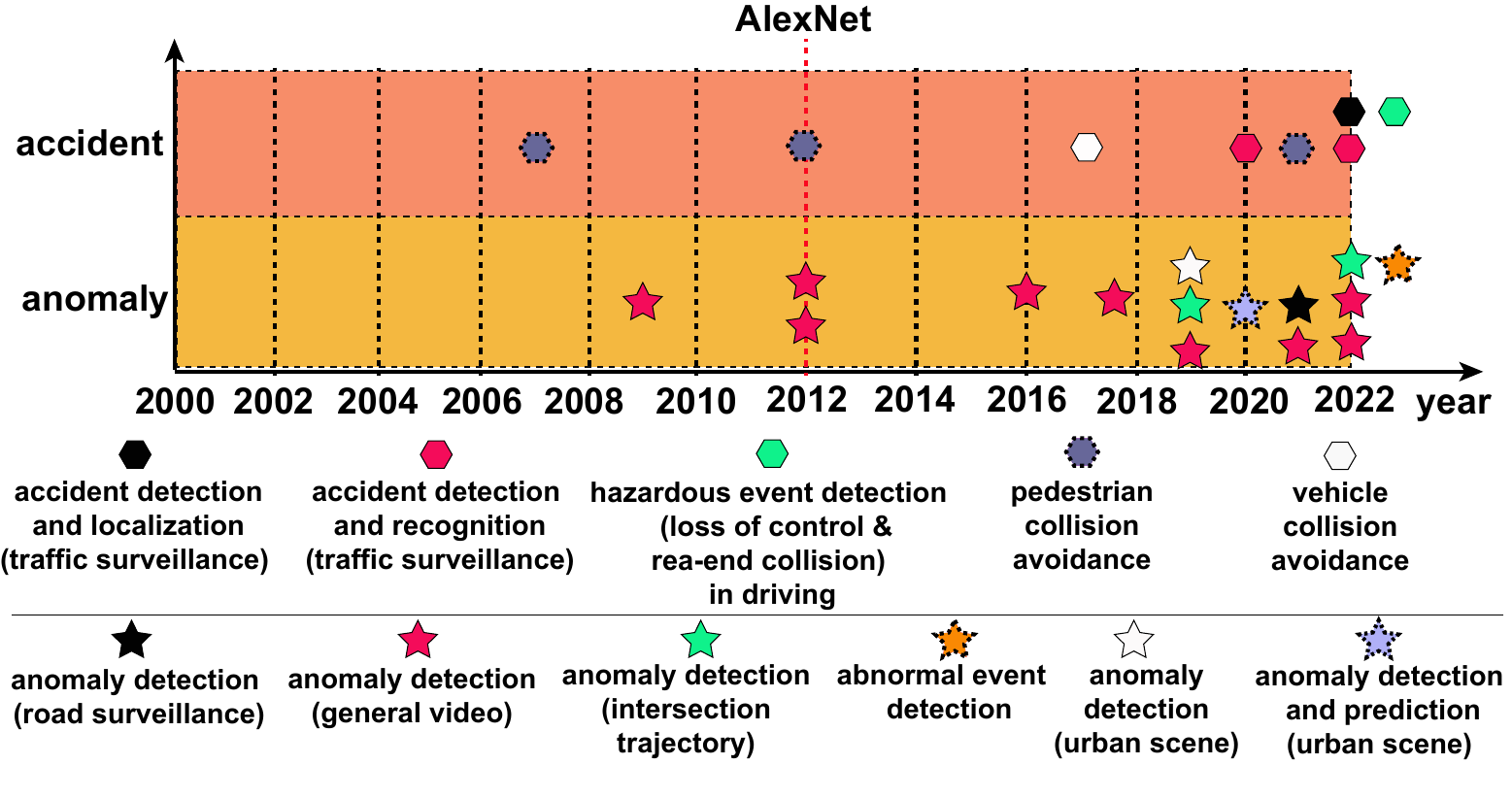}
	\caption{The evolution process of 24 related surveys. Notably, each node in this figure represents a high-quality survey detailed in Table. \ref{tab1}.}
	\label{fig3}
\end{figure}

\begin{figure*}[!t]
	\centering
	\includegraphics[width=\hsize]{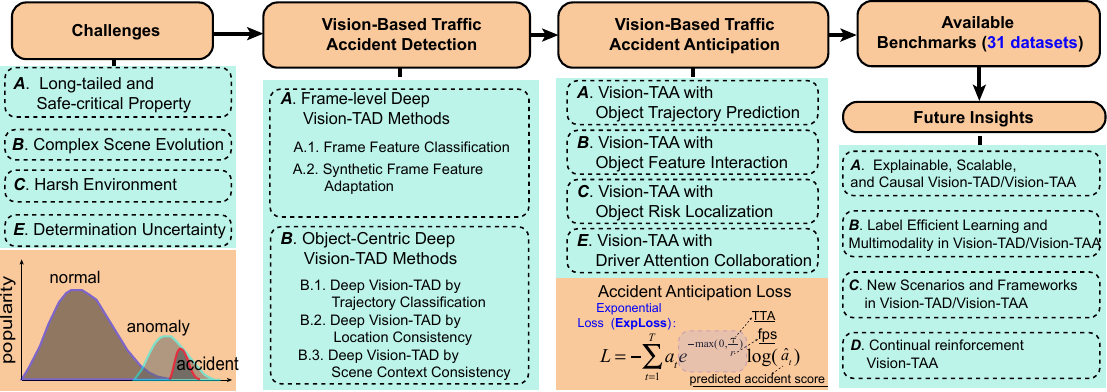}
	\caption{The survey taxonomy in this work. We first review the challenges in Vision-TAD and Vision-TAA. Then, we summarize the approaches for Vision-TAD and Vision-TAA in different granularities, where the main inputs, accident detection criteria, and anticipation loss are presented. The available benchmarks and metrics are reviewed clearly, and the open problems and future insights are finally discussed.}
	\label{fig4}
\end{figure*}
\begin{table}[!t]\footnotesize
 \centering
 \caption{The surveys of traffic vision safety.}
 \renewcommand{\arraystretch}{1}
 \setlength{\tabcolsep}{0.3mm}{
  \begin{tabular}{c|c|c}
\toprule[0.8pt]
   Situation & Topics & Survey Ref. \\
   \hline\hline
   Anomaly&\makecell{\rule{0pt}{15pt}anomaly detection \\in generic situations\\\rule{0pt}{20pt}  anomaly detection \\in road traffic \\scenes (urban scene, \\intersection trajectories)}&\makecell{\cite{chandola2009anomaly,DBLP:journals/tsmc/OluwatoyinW12,DBLP:journals/tsmc/SodemannRB12,DBLP:conf/ised/PatilB16,DBLP:journals/jaciii/MaDH17},\\\cite{DBLP:journals/tcsv/ZhangNH0Y21,DBLP:journals/ivc/NayakPD21,chandrakala2022anomaly,DBLP:journals/pami/RamachandraJV22,DBLP:journals/corr/abs-2302-05087},\\\cite{DBLP:journals/csur/TranVVNN23,DBLP:journals/corr/abs-2302-05087,jiang2023weakly}\\\rule{0pt}{25pt} \cite{DBLP:journals/csur/KumaranDR21,DBLP:journals/access/DjenouriBLDC19,DBLP:journals/tbd/ZhangLYLHZ22}, \cite{minnikhanov2022evaluation}}\\
   \hline\hline
   Accident &\makecell{\rule{0pt}{15pt}accident analysis\\\rule{0pt}{10pt}accident recognition\\\rule{0pt}{10pt}pedestrian or vehicle \\collision avoidance}&\makecell{\rule{0pt}{10pt}\cite{DBLP:journals/corr/abs-2203-10939,DBLP:journals/corr/abs-2208-09588}\\\rule{0pt}{10pt}\cite{evans2020evolution,DBLP:phd/hal/Maaloul18a,DBLP:journals/corr/abs-2208-09588}\\\rule{0pt}{15pt}\cite{DBLP:journals/tits/GandhiT07,llorca2012stereo,DBLP:journals/tits/BilaSKA17,DBLP:journals/tits/FuLYL022}}\\
   \hline
 \end{tabular}}
 \label{tab1}
\end{table}

In order to address the above challenges, the dataset platforms and inference models in this field have developed with large progress. This survey aims to summarize the developing process comprehensively and looks for possible new insights for Vision-TAD and Vision-TAA. In addition, leveraging the traffic anomaly research, this survey also wants to clarify the research paradigm difference and search for more targeted solutions for Vision-TAD and Vision-TAA.

\subsection{Distinction from Other Reviews} 

To be clear, we extensively investigate the related surveys and summarize the evolution process from the anomaly and accident aspects in Fig. \ref{fig3} and Table. \ref{tab1}. We can see that the surveys for traffic accident understanding are rather fewer than the ones for anomaly detection. Video anomaly detection in general surveillance systems (\emph{e.g.}, shopping malls, subways, etc.) takes the largest part from traditional \cite{chandola2009anomaly,DBLP:journals/tsmc/OluwatoyinW12,DBLP:journals/tsmc/SodemannRB12} to deep learning approaches \cite{DBLP:journals/csur/KumaranDR21,DBLP:journals/access/DjenouriBLDC19,DBLP:journals/tbd/ZhangLYLHZ22,minnikhanov2022evaluation}. This fact makes the video anomaly detection infer the appearance or motion features in general videos but the specific scene rules, such as traffic rules, are not exploited well. Compared with video anomaly detection \cite{DBLP:journals/tcsv/PiciarelliMF08,DBLP:journals/tcsv/ZhangNH0Y21,DBLP:journals/tcsv/WuZSWW23,DBLP:journals/tcsv/ZhongCHTR22,xing2023visual}, vision-based traffic accident perception shows the most inadequate research mainly due to the collection difficulty of accident video data. Most relating to this work, various surveys on traffic accident situations concentrate on traffic accident recognition or collision avoidance from specific occasions (\emph{e.g.,} intersection, urban scene), road participants (\emph{e.g.,} vehicles-centric and pedestrian-centric), and applications (\emph{e.g.,} surveillance safety \cite{DBLP:journals/corr/abs-2208-09588,evans2020evolution} and autonomous driving \cite{DBLP:journals/tits/BilaSKA17,DBLP:journals/tits/FuLYL022}). However, the survey for vision-based traffic accident detection is few-explored. Hence, the \textbf{distinctions} to other reviews are:
\begin{itemize}
\item This work presents a focused review of Vision-TAD with the detailed method formulation, datasets, and issues in the deep-learning era. 
\item This work forms the first Vision-TAA investigation from the primary research pipelines, issues, and benchmarks.
\end{itemize}

\subsection{Taxonomy and Contributions}
 Based on the investigation, we organize the taxonomy of this survey in Fig. \ref{fig4}. Based on the challenges, we will first present the method details of Vision-TAD and Vision-TAA, where the advantages and disadvantages are discussed in detail. After that, we reveal the public datasets and evaluation metrics to make this survey useful for future research. Finally, we discuss the insights for future research.  To summarize, the \textbf{contributions} of this survey are threefold.
 \begin{itemize}
\item We present the first review for vision-based traffic accident detection in deep learning and present the first-ever review for vision-based accident anticipation. We clarify the difference between video anomaly detection and traffic accident detection in this survey.
 
\item We analyze the pros and cons of the research prototypes of Vision-TAD and Vision-TAA and the details of public benchmarks are provided from the dataset attributes, downloading links, and evaluation metrics.
 
\item Promising insights and trends for future research are presented, including the explainable, scalable, causal properties, label efficient learning, multimodality complementary, new scenarios and frameworks, and the continual reinforcement Vision-TAA.
  \end{itemize}

\section{Vision-Based Traffic Accident Detection}
Vision-TAD has a close relation to the traffic video anomaly detection task \cite{DBLP:journals/tcsv/ZhangNH0Y21}. Differently,  traffic accidents involve collision, crash, and incident situations, while video anomaly contains a wider scope than accident situations. Through the development of two decades, the methods for Vision-TAD undergo from classic model-driven methods to data-driven deep learning approaches. The core problem in Vision-TAD is to extract the robust appearance or motion features for the representation of video frames, spatial-temporal video volumes, or trajectories \cite{akoz2010video,ghahremannezhad2020real,datondji2016survey,essam2022detection}. Actually, if the accident label for video frames or locations is not used, the classic pipelines for traffic accident detection have little difference from video anomaly detection. To be focused on this work, we will not describe the traffic anomaly detection approaches, and the related works can be reviewed in previous surveys \cite{DBLP:journals/pami/RamachandraJV22,DBLP:journals/csur/TranVVNN23}. In this work, we mainly review how to detect traffic accidents in data-driven deep-learning models. Based on the model architectures, we divide the methods into the categories of frame-level deep Vision-TAD and object-centric deep Vision-TAD. 

\subsection{Frame-level Deep Vision-TAD Methods}

Frame-level deep Vision-TAD methods aim to detect video accidents in traffic scenes with the frame-level feature representation or accident determination. In addition, because it is hard to annotate enough accident data in practice, synthetic accident video generation is also attractive in this field.

\subsubsection{Frame Feature Classification}

Traffic accidents are commonly contained in traffic anomalies. In order to find the true accident from the chaotic anomaly set, some works directly annotate the accident window labels \cite{DBLP:conf/eccv/YouH20,DBLP:journals/bdcc/AhmedZASAASARK23,DBLP:conf/idsta/SrinivasanSIN20}. For example, You and Han \cite{DBLP:conf/eccv/YouH20} present a traffic understanding dataset, where they localize the accident window by classifying the feature embedding of video frames by the 3D-CNN model. Based on the corner case property, the distance between the frames with frame feature centers determines the accident score. In this category, the reconstruction error similar to video anomaly detection is a common choice. A common choice of reconstruction error is Frobenius norm \cite{DBLP:journals/tits/SinghM19}, defined as
\begin{equation}
	L=\frac{1}{T}\sum_{t=1}^{T} {||x_t-\tilde{x_t}||_\mathcal{F}},
\end{equation}
where $x_t$ denotes input feature of $t^{th}$ frame, $\tilde{x_t}$ is the reconstruction of $x_t$, and $T$ is the number of frames. Frame-level reconstruction  \cite{DBLP:journals/tits/SinghM19,DBLP:journals/ict-express/PawarA22} mainly takes the encoder-decoder architectures to learn the appearance feature representation.
 
In addition to the direct classification approaches with accident labels,  some works explore the \textbf{coarse-to-fine} framework to find the traffic accident from the traffic anomaly, where the coarse anomaly frames are obtained by clustering the frame features and then refining the detection of accident frames by object-level confirmation with accident frame classification \cite{zhou2022spatio} or accident-normal pattern cross-validation \cite{DBLP:journals/tits/VijayDCNK23}. 

The aforementioned methods for Vision-TAD focus on the video frame feature classification or reconstruction. Commonly, these kinds of approaches are easily constrained by the available video data because traffic accident often occurs in a very short duration in traffic scenes. Therefore, it is hard to distinguish the accident from the abnormal frames because of the sample imbalance. 

\subsubsection{Synthetic Frame Feature Adaptation}
Due to the collection difficulty, some works leverage synthetic video data to assist accident detection in real scenarios via the domain adaption approaches. Based on different observation views, such as surveillance and dashcam views, the attributes of the synthetic data are different. For example, taking the monocular video frame as a background reference, Batanina \emph{et al.} \cite{DBLP:conf/ipta/BataninaBKKB19}  generate accident videos by GTA-V game engine, and utilize Triple Loss Network (TripNet) \cite{DBLP:journals/algorithms/BekkouchYG0K19} to fulfill an unsupervised synthetic-to-real domain adaption. Following this insight,  Tamagusko \emph{et al.} \cite{Tamagusko2022} synthesizes accident cases by making a collision of 3D virtual vehicles. Then, the pre-trained binary CNN model on synthetic video frames (accident and accident-free frame annotation) is fine-tuned on the real video frames. Because of the diversity of synthetic data, Luo and Han \cite{luoICASSP2023} adopt the accident frame classifier trained on synthetic videos to the real videos directly. 

Besides the single view observation in synthetic frame generation, Vijay \emph{et al.} \cite{DBLP:journals/tits/VijayDCNK23} generate 400 accident videos, while each accident event is recorded by multiple perspectives for a full-range observation. Under this complete observation with multiple views, the spatial-temporal features of multi-view video groups are extracted by two-branch DCNNs and classified in \emph{softmax} function. 

\textbf{Trends:} Recent generation models, such as diffusion models \cite{croitoru2022diffusion,ho2022video}, show a powerful ability for realistic and controllable video or image generation constrained by some conditions. In addition, traffic accident has a clear definition that they cause a collision. Therefore, involving traffic scene knowledge (text description) to fulfill flexible video generation and explainable Vision-TAD works is promising. For instance, Wang \emph{et al.} \cite{wang2023towards} present a survey on explainable visual anomaly detection methods in visual anomaly detection, which considers the attention-based, input-perturbation-based, reasoning-based, and intrinsically interpretable visual anomaly detection methods. Although it does not focus on the traffic scenes, the insights are helpful for traffic accident detection. 

In addition, the frame-level deep Vision-TAD methods need to encode the frame background, which may involve many influencing features from the background regions. Therefore, the object-centric deep Vision-TAD methods can eliminate the background and focus on the key object in reasoning.

\subsection{Object-Centric Deep Vision-TAD Methods}
Compared with frame-level deep Vision-TAD methods, object-centric approaches concentrate on object-level temporal consistency and follow the object detection and tracking stages to generate the trajectories, where various detectors (\emph{e.g.,} mask-RCNN \cite{chand2020computer}, FasterRCNN \cite{lee2019application}, YOLOv4 \cite{ghahremannezhad2022real}, YOLOv5 \cite{xia2022research}, \emph{etc.}) and many object association approaches (SORT \cite{DBLP:conf/icip/BewleyGORU16}, DeepSort \cite{DBLP:conf/icip/WojkeBP17}, Kalman Filter \cite{ghahremannezhad2022real,moayed2019surveillance}, Hungarian algorithm \cite{ghahremannezhad2022real}, \emph{etc.}) are utilized. Object-centric Vision-TAD methods can get rid of the influence of the complex background. However, removing the background information has a double-edged sword. They introduce the perception error in the pre-detection and tracking steps, and the accident inference cannot be achieved by the end-to-end framework but is universal in frame-level deep Vision-TAD models. Therefore, the core issue is to find the true accident trajectories from the clutter trajectory points of the object group. Based on the investigation, the object-centric deep Vision-TAD methods can be divided into trajectory classification, location consistency, and scene context consistency.

\subsubsection{Deep Vision-TAD by Trajectory Classification}

Before the accident determination, trajectory features are encoded by various kinemetric clues (\emph{e.g.,} the velocity and acceleration \cite{chand2020computer}) or deep features. Then, common formulations learn the dominant trajectory feature clusters and classify the trajectory feature set for finding accident ones, where the accident is determined by the distance between the trajectory feature with the learned cluster centers. 
In the training stage of the classifiers, the core problem is to find the potential labels of normal and accidental trajectories. Generally, annotating all the normal and accidental trajectories is difficult, so some easily annotated trajectories can be treated as a start-up with semi-supervised learning.  
For example, Santhosh \emph{et al.} \cite{santhosh2021vehicular} generate the pseudo-label of normal trajectories and accident trajectories by t-Distributed Stochastic Neighbor Embedding (t-SNE) in the training stage, and then the CNN and Variational Autoencoder (VAE) are adopted to classify the trajectory features. Chakraborty \emph{et al.} \cite{chakraborty2018freeway} firstly obtain the trajectories of the road participants by YOLO-v3 \cite{redmon2018yolov3} and Simple Online Realtime Tracking (SORT)  \cite{DBLP:conf/icip/BewleyGORU16}. Then, they estimate the distribution of the labeled trajectories by Maximum Likelihood (ML) estimation. Classification-based object-centric Vision-TAD needs to annotate the accident labels of the trajectories, which is difficult and owns the data bias for the few-shot accident trajectories. Therefore, some works directly introduce the physical kinematic clues, such as velocity, acceleration, and direction of trajectories, to determine the accident.

 \begin{figure}[!t]
 \includegraphics[width=\hsize]{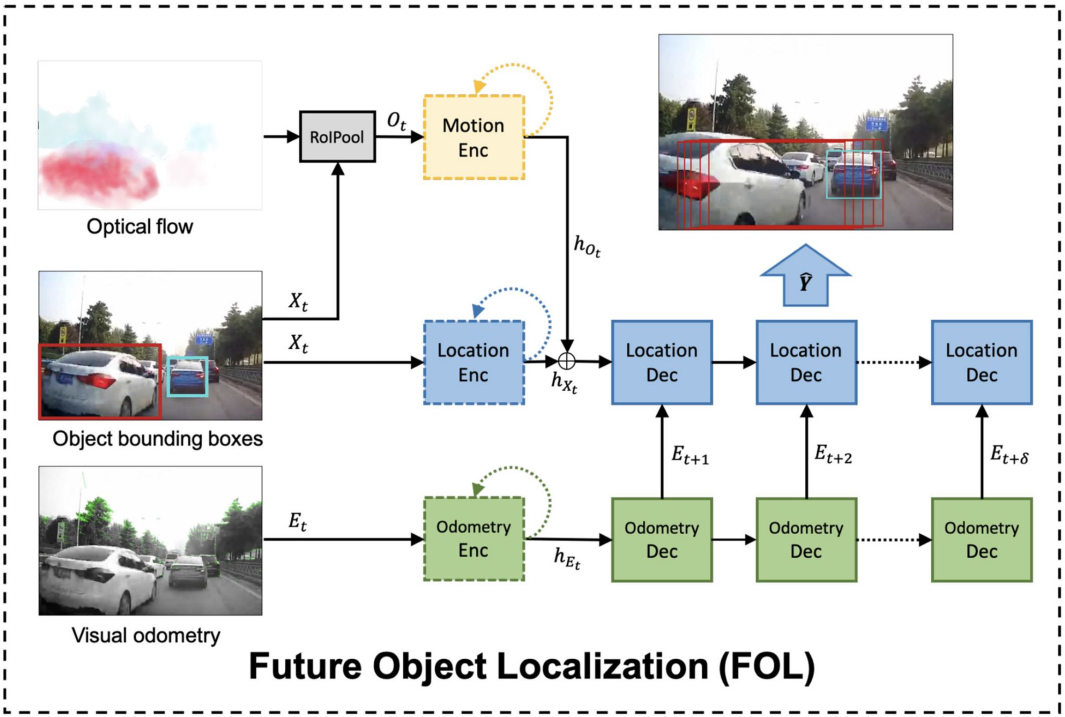}
  \caption{The flowchart of Future Object Localization (FOL) in \cite{yao2022dota}.}
  \label{fig5}
\end{figure}

\subsubsection{Deep Vision-TAD by Location Consistency}
\label{loc}
Location consistency measurement in deep Vision-TAD works involves severe occlusion and scale change. Le \emph{et al.} \cite{le2020attention} annotate a new dataset for object detection in accident scenarios, where the collision is estimated by the overlapping of the detected bounding boxes by an Attention-RCNN detector.

Some works check the location consistency further with the Standard Deviation (STD) \cite{DBLP:conf/iros/YaoXWCA19,yao2022dota} of object location sequence in the time window. 
For example, the work \cite{taccari2018classification} uses a YOLO detector to detect the object, and the dense optical flow of images is generated by Farneback’s algorithm \cite{farneback2003two}.  Then the optical flow feature within the associated object bounding boxes in consecutive frames is extracted by computing the \emph{max}, \emph{min}, \emph{standard}, and \emph{average} motion offset. These features are then fed into a random forest to classify the location motion features.  Recently, Yao \emph{et al.} \cite{DBLP:conf/iros/YaoXWCA19,yao2022dota} propose an unsupervised traffic accident detection method, which models the Future Object Localization (FOL) by predicting the object locations with the center ($c^x, c^y$), the width $w$ and the height $h$ in the next $T$ future frames, where the optical flow is also used to guide the location prediction with the Gated Recurrent Unit (GRU) model \cite{dey2017gate}, as shown in Fig. \ref{fig5}. In addition, they introduce visual odometry in the location prediction for considering the ego-motion of cars. Considering these multiple motion features, the location prediction commonly is very short with about 0.5 seconds. The traffic accident is determined by checking the Standard Deviation (STD) and Intersection over Union (IoU), and are defined as
\begin{equation}
 S_1(t)= 1-\frac{1}{N}\sum\limits_{i=1}^{N}\text{IoU}((\frac{1}{T}\sum\limits_{m=1}^{T}\hat{Y}_{t,t-m}^{i}),Y_{t,0}^{i})
\end{equation}
where $\text{IoU} = (A\cap B)/(A\cup B)$, and
\begin{equation}
 S_2(t)= \frac{1}{N}\sum\limits_{i=1}^{N}\max\limits_{\{c^x,c^y,w,h\}}\text{STD}(\hat{Y}_{t,t-m})
\end{equation}
The parameter $N$ is the total number of observed objects, $A$ and $B$ denote the bounding box region of actual and predicted one, $\hat{Y}_{t,t-m}^{i}$ is the predicted bounding box from time $t-m$ to $t$ of the object $i$.
The aforementioned object-centric deep Vision-TAD works mainly treat the trajectories as separate indicators for subsequent accident determination. The traffic accident has an apparent feature of scene context inconsistency. 
  \begin{table*}[!t]\footnotesize
  \centering
  \caption{The characteristics of representative Vision-TAD works, where the Years, Models, Inputs, Accident Determination CriterIa (Accident Deter. Criteria), pre-steps (\emph{i.e.,} detection (Det.) and Tracking (Track.)), Accident Occasions (Accd. O), and \emph{Supervising Mode}, which consists of the supervised (SuperV), semi-supervised (SemiS), weakly-supervised (WeakS), self-supervised (SelfS), and unsupervised (USuper) modes.}
\renewcommand{\arraystretch}{1.2}
     \setlength{\tabcolsep}{0.1mm}{
\begin{tabular}{c|c|c|c|c|c|c|c}
\toprule[0.8pt]
Ref. &Years&Models&Inputs&Accident Deter. Criteria&Pre-steps& Accd. O&Super. M\\
       \hline\hline
Kamijo \emph{et al.} \cite{DBLP:journals/tits/KamijoMIS00}&2000&HMM&Gray frames&object distance&Det. Track&intersection&SuperV\\
\hline
Fatih and Li \cite{porikli2004traffic}&2004&HMM&RGB frames&maximum likelihood&Patch Parsing&highway&SuperV\\
\hline
Zou \emph{et al.} \cite{zou2011traffic}&2011&HMM+one class SVM&RGB frames&pattern bias&Det. Track&intersection&SuperV\\
\hline
Chakraborty \emph{et al.} \cite{chakraborty2018freeway}&2018&Contrastive Likelihood&RGB frames&maximum likelihood&YOLO-v3, Track&highway&SemiS\\
\hline
Singh and Mohan \cite{DBLP:journals/tits/SinghM19}&2019&Stacked Autoencoder&RGB frames&reconstruction error&YOLOv5, Track&urban city&USuper\\
\hline
Yao \emph{et al.} \cite{DBLP:conf/iros/YaoXWCA19}&2019&CNN+GRU&\makecell{object bounding boxes\\optical flow\\ego-vehicle’s pose}&IOU deviation&Mask-RCNN, Track&dashcam&USuper\\
\hline
Moayed \emph{et al.} \cite{moayed2019surveillance}&2019&Step-by-Step&RGB frames&IOU&YOLO-v2, KF&highway&-\\
\hline
Batanina \emph{et al.} \cite{DBLP:conf/ipta/BataninaBKKB19}&2019&CNN+Domain Adaptation&RGB frames&classification score&-&highway&SuperV\\
\hline
Huang \emph{et al.} \cite{DBLP:journals/tsas/HuangHRR20}&2020&CNN&RGB frames&traj. intersection check&YOLO&urban city&SuperV\\
\hline
Le \emph{et al.} \cite{le2020attention}&2020&Attention RCNN&RGB frames&Box regression loss&FasterRCNN&dashcam&SuperV\\
\hline
Roy \emph{et al.} \cite{roy2020detection}&2020&Siamese Interaction LSTM& vehicle bounding boxes
&collision energy&Det. Track&intersection&SuperV\\
\hline
Haresh \emph{et al.} \cite{DBLP:conf/ivs/HareshKZT20}&2020&CNN Encoder-Decoder&RGB frames&reconstruction error&Det. (RPN)&dashcam&USuper\\
\hline
 Nguyen \emph{et al.} \cite{DBLP:conf/mir/NguyenDDT20}&2020&GAN&\makecell{RGB frames\\the stacked motion frames}&reconstruction error&-&urban city&SelfS\\
 \hline
Santhosh \emph{et al.} \cite{santhosh2021vehicular}&2021&CNN+VAE&Trajectories&reconstruction error&Context tracker&urban city&USuper\\
\hline
 Hu \emph{et al.} \cite{DBLP:journals/eswa/HuWCG21}&2021&Trajectory CNN+LSTM&Trajectories&softmax, risk score&Det. Track&highway&SemiS\\
 \hline
  Zhou \emph{et al.} \cite{zhou2022spatio}&2022&CNN+one-class SVM&RGB frames&classification score&Faster-RCNN&dashcam&USuper\\
  \hline
 Fang \emph{et al.} \cite{fang2022traffic}&2022&GRU, LSTM&\makecell{RGB frames\\object bounding boxes \\optical flow}&MSE+IOU&Mask-RCNN+DeepSort&dashcam&SelfS\\
 \hline
 Hajri and Fradi \cite{DBLP:conf/avss/HajriF22}&2022&Transformer&RGB frames&softmax&-&dashcam&SelfS\\
 \hline
 Yao \emph{et al.} \cite{yao2022dota}&2022&CNN+GRU&\makecell{object bounding boxes\\optical flow\\visual odometry}&standard IOU deviation&Mask-RCNN, Track&dashcam&USuper\\
  \hline
 Vijay \emph{et al.} \cite{DBLP:journals/tits/VijayDCNK23}&2023&Two-branch DCNN&\makecell{Synthetic and \\Real RGB frames}&classification score&-&urban city&SuperV\\
  \hline
  \end{tabular}}
  \label{tab2}
  \end{table*}
  
\subsubsection{Deep Vision-TAD by Scene Context Consistency}

Scene context consistency usually models the interaction graph of relations among the detected objects in traffic scenes, and follows the assumption that normal situations have a stable scene context evolution but a sudden change of scene context appears in accident situations. The work \cite{DBLP:journals/ieicetd/YamamotoKT22} models a near-miss classification method in the videos captured by the event camera. They define four kinds of risk states \{\emph{high, middle, low, bit, and no near-miss}\}, and achieve the near-miss event classification by multi-task learning of temporal frame and object interaction features. Fang \emph{et al.} \cite{fang2022traffic} model the location consistency, appearance consistency, and scene context consistency together to infer the traffic accident detection in dashcam videos, where a graph-based generative adversarial network is modeled to evaluate the scene context consistency. Because the interaction of a certain participant may involve many surrounding targets and the key relation may be a sub-link in the redundant interactions, Roy \emph{et al.} 
\cite{DBLP:journals/tits/RoyIMF22} propose a Siamese Interaction LSTM (SILSTM) to involve interactive temporal attention on the hidden states of objects. The accident is determined by a collision energy model \cite{robicquet2016learning}, defined as
\begin{equation}\small
E_c({\mathbf{v}};s_i,{\mathbf{s}}_{j\neq i}|\sigma_d,\sigma_w,\beta)=\sum_{j\neq i}w(s_i,s_j)\exp(\frac{d^2({\bf{v}},s_i,s_j)}{2\sigma_d^2}),
\label{eq:4}
\end{equation}
where 
\begin{equation}
w(s_i,s_j)=\exp(-\frac{|\Delta{\mathbf{p}}_{ij}|}{2\sigma_w})\cdot(\frac{1}{2}(1-\frac{\Delta{\mathbf{p}}_{ij}}{|\Delta{\mathbf{p}}_{ij}|}\frac{{\mathbf{v}}_i}{|{\mathbf{v}}_i|}))^\beta,
\end{equation}
and 
\begin{equation}
d^2({\bf{v}},s_i,s_j)=|\Delta{\mathbf{p}}_{ij}-\frac{\Delta{\mathbf{p}}_{ij}({\mathbf{v}}-{\mathbf{v}}_j)}{|{\mathbf{v}}-{\mathbf{v}}_j|^2}{(\mathbf{v}}-{\mathbf{v}}_j)|.
\end{equation}
The parameter $\sigma_d$ is the safe distance that a vehicle maintains with surrounding objects. $\sigma_w$ is the distance that a vehicle (with the position and velocity state $s_i=\{{\mathbf{p}}_i, {\mathbf{v}}_i\}$) can prevent the collision in overtaking, merging, and other behaviors. $\beta$ denotes the largest value of the weighting function for turning distance, and $\Delta {\mathbf{p}}_{ij}$ specifies the distance between vehicle $i$ and target $j$. The collision-free goal is to minimize Eq. \ref{eq:4} and solve $\sigma_d,\sigma_w,\beta$ for certain vehicle $i$. Traffic accident detection is formulated as the detection of high collision energy points.

The object-centric deep Vision-TAD follows the object detection and tracking framework, and some works adopt location prediction or trajectory prediction in traffic accident determination. This architecture can focus on the road object while introducing perception error in each stage. In the following, we present a discussion of Vision-TAD works.

\subsection{Discussions}
\label{tad:tadc}
We present some representative Vision-TAD works in Table. \ref{tab2} from the aspects of publication years, models, inputs, accident determination criteria, pre-steps, accident occasions, and supervising mode. Accident detection in surveillance view, such as urban city and highway scenarios, takes a larger proportion than dashcam view which has attracted more attention in recent years. From this table, we can see that the Hidden Markov Model (HMM) is focused early on classic accident detection models, and supervised accident detection takes the main part. Autoencoder is the common architecture for unsupervised accident detection and unsupervised learning becomes the attractive learning mode because of the difficult accident data annotation. Few research efforts notice the Transformer, which will be focused on in the future. Besides, because of the long-tailed distribution for accidents, the synthetic data generation also can be concentrated with large efforts. Most of the representative works belong to the object-centric category because this formulation can get rid of the influence of video background. 

The dominant evaluation metric for accident determination is the reconstruction error, which is mainly evaluated by the distance between the learned normal frame feature in traffic scene videos with the testing data. The IOU deviation in adjacent frames shows an apparent clue for scene stability, which is promising for object-centric accident detection approaches. However, most of these works need to pre-detect and track the object in advance for object-centric accident detection. This kind of step-by-step formulation is easily influenced by severe environment conditions. Although there are some works that annotate the spatial location of objects \cite{yao2022dota}. The benchmark scale is rather limited compared with other ones in normal situations.
In addition, based on the investigation, we find that few works have been exploited for the distribution difference inference of general traffic anomaly and traffic accidents. In fact, it is hard to give a clear separation, but scene knowledge may be promising \cite{DBLP:journals/tits/YuMMKF22,DBLP:journals/corr/abs-2212-09381}.

\section{Vision-Based Traffic Accident Anticipation}
Compared with Vision-TAD, Vision-TAA can actively predict the potential collision early, so as to obtain enough time for safe decisions and collision avoidance. Actually, most works for Vision-TAA belong to the safe driving field. In addition, different from Vision-TAD, Vision-TAA is more challenging because of the future uncertainty and the complex driving scenes. Based on the investigation, we will review the Vision-TAA works from the following categories. 

\subsection{Vision-TAA with Object Trajectory Prediction}
Early research pipelines in Vision-TAA with object location association borrow the trajectory prediction models and follow the pipeline of \emph{detection-tracking-trajectory prediction-accident determination} \cite{DBLP:conf/ciss/HarisMRF21}. In the meantime, early accident prediction research focuses on the surveillance fields, and the works follow the direct physical distance computation for accident prediction on the predicted trajectories. 
For example, Hu \emph{et al.} \cite{hu2003traffic} model road collision prediction as a trajectory segment intersection point detection problem at each time. If the distance between the intersection points is smaller than the diagonal length of the vehicles, a collision will appear. Because this direct distance measurement may introduce false alarms, they further introduce the self-organizing neural network to learn the trajectory pattern and predict the accident by locating each pattern-inconsistent trajectory \cite{DBLP:journals/tvt/HuXXTM04}. Different from the location consistency in Sec. \ref{loc}, the trajectory prediction here has a longer future time window for maximizing the Time-to-Accident (TTA) interval. 

Based on the regulated traffic rules of highway scenarios, some works \cite{xiong2017new,gutierrez2020modern} adopt some classification models, such as SVM, HMM, decision trees, \emph{etc.}, to recognize the future accidents by the vehicle trajectories. Shan \emph{et al.} \cite{DBLP:journals/mms/ShanZZ17} analyze the safe distance within the future trajectories predicted by Autoregressive Integrated Moving Average (ARIMA) \cite{lee1999application} under a linear motion assumption for each vehicle. In addition to the distance measurement for each time, the distance between the subsequent predicted trajectory segment with the current predicted segment is also a clue for accident prediction \cite{DBLP:journals/corr/abs-2101-08463}, \emph{i.e.,} evaluating the standard deviation of the several trajectory segments in the prediction window.

\subsection{Vision-TAA with Object Feature Interaction}
\begin{figure}[!t]
  \centering
 \includegraphics[width=\hsize]{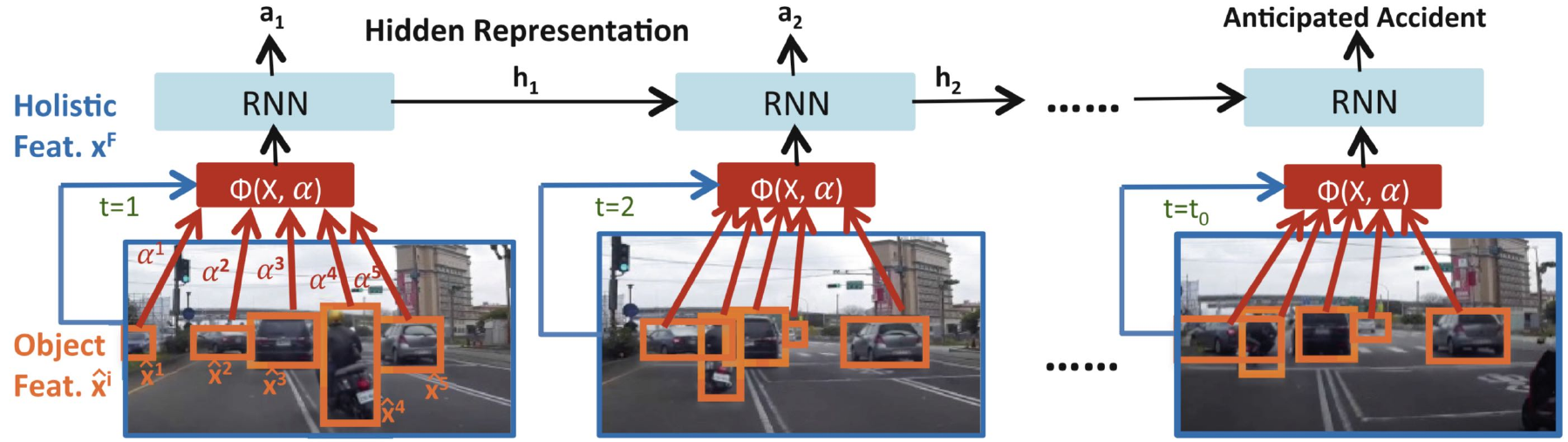}
  \caption{The flowchart of accident anticipation in \cite{DBLP:conf/accv/ChanCXS16}.}
  \label{fig6}
\end{figure}

In recent years, traffic accident anticipation in dashcam videos become popular because of the development of autonomous driving systems. However, different from the surveillance field, the frequent occlusion of road participants makes the aforementioned physical distance measurement infeasible. Therefore, some works begin to explore the interactive relation within the participant group. Borrowing the insights from event anticipation in robotic community \cite{DBLP:journals/pami/KoppulaS16,DBLP:conf/iros/MainpriceB13}, Chan \emph{et al.} \cite{DBLP:conf/accv/ChanCXS16} propose the first accident anticipation work in dashcam videos using deep learning models.  They model a Dynamic-Spatial-Attention (DSA) \cite{DBLP:conf/accv/ChanCXS16} and Recurrent Neural Network (RNN) \cite{DBLP:journals/corr/ZarembaSV14} to correlate the interaction among the objects in road scenes, as shown in Fig. \ref{fig6}. Many accident anticipation works are inspired by \cite{DBLP:conf/accv/ChanCXS16} with different temporal or spatial interaction modeling.

For the object interaction, the work \cite{DBLP:conf/accv/ChanCXS16} firstly uses Faster-RCNN \cite{DBLP:journals/pami/RenHG017} to generate less than 20 object regions in each frame, and the motion feature of each object is obtained by the median motion offset in each region. They also combine the frame appearance feature and frame clip motion feature into the observations encoded by the VGG \cite{DBLP:journals/corr/SimonyanZ14a} and Improved Dense Trajectory (IDT) \cite{wang2013action}, respectively. 
 Inspired by this work, Karim \emph{et al.} \cite{karim2022dynamic} propose a Dynamic Spatial-temporal Attention (DSTA) network for traffic accident anticipation. They design the Dynamic Temporal Attention (DTA) for weighting the hidden states of past $N$ frames ${\bf{H}}_{t-1}=[{\bf{h}}_{t-1},...,{\bf{h}}_{t-N}]$, and defined as
\begin{equation}
{\bf{h}}_t^{'}=[{\bf{\delta}}_{t-1},{\bf{H}}_{t-1}]_r,
\end{equation}
where ${\bf{\delta}}_{t-1}$ is the attention weights and computed by $\gamma({\bf{W}}_{dta}\tanh({\bf{H}}_{t-1}))$. ${\bf{W}}_{dta}$ is the weight parameters of DTA, $[.,.]_r$ denotes the row-wise inner product, and $\gamma$ is the \emph{softmax} operator.

\begin{figure}[!t]
	\centering
	\includegraphics[width=0.7\hsize]{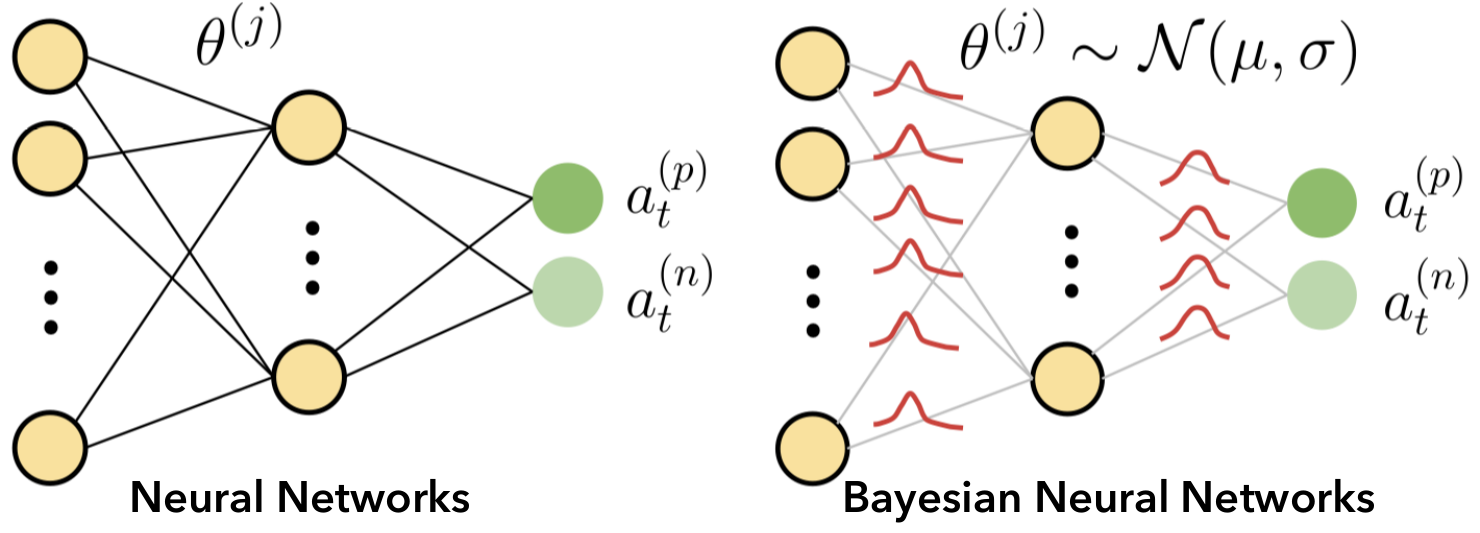}
	\caption{The structure comparison of neural networks and Bayesian Neural Networks (BNN) with a parameter sampling from a Gaussian distribution. This figure is from \cite{bao2020uncertainty}.}
	\label{fig7}
\end{figure}

With the dynamic spatial and temporal attention model, the work \cite{karim2022dynamic} formulates a Spatial-Temporal Relational Learning (STRL) model with GRU, which is inspired by the work \cite{bao2020uncertainty} that takes the graph convolutional recurrent network (GCRN) \cite{seo2018structured} to model STRL. STRL in \cite{karim2022dynamic} is fulfilled by inputting ${\bf{h}}_{t-1}^{'}$ and the feature observation ${\bf{X}}_t$ into GRU is defined as 
\begin{equation}
{\bf{h}}_{t}=\text{GRU}({\bf{h}}_{t-1}^{'},{\bf{X}}_t, {\bf{W}}_{gru}),
\end{equation}
where ${\bf{W}}_{gru}$ is the weights of GRU. As for the work \cite{bao2020uncertainty}, because the number of the detected objects in video frames may be different, STRL is designed as
\begin{equation}
\begin{aligned}
{\bf{Z}}_{t}=\text{GCN}([\text{GCN}({\bf{X}}_{t}, {\bf{A}}_{t}),{\bf{h}}_{t}],{\bf{A}}_{t}),\\
{\bf{h}}_{t+1}=\text{GCRN}({\bf{h}}_{t},[{\bf{X}}_{t}, {\bf{Z}}_{t}]),
\end{aligned}
\end{equation}
where the spatial relation is fulfilled by two layers of graph convolutional network (GCN), and ${\bf{A}}_{t}$ is the affinity matrix computed on ${\bf{X}}_{t}$. In addition, the work \cite{bao2020uncertainty} also considers the anticipation uncertainty and determines the accident score by a Bayesian Neural Network (BNN) \cite{gal2016dropout}, as shown in Fig. \ref{fig7}. Similarly, Wang \emph{et al.} \cite{wang2023gsc} also borrow the GCN to model the interaction of objects in driving scenes, where a spatial-temporal continuity is learned for the edge weight updating in accident anticipation.

The frame-level accident score $a_t$ for each time can be computed by
\begin{equation}
	a_t=\text{softmax}(\phi(\phi({\bf{h}}_t;\theta_1);\theta_2)),
	\label{eq:5}
\end{equation}
where $\phi$ denotes the fully-connected layer, $\theta_1$ and $\theta_2$ are the parameters of the two-fully-connected layers.

\subsection{Vision-TAA with Object Risk Localization}
Traffic accidents certainly involve a safe risk for further movement, and risk localization aims to localize the risk regions or objects for early accident prediction as much as possible. Zeng \emph{et al.} \cite{DBLP:conf/cvpr/ZengCCNS17} propose an agent-centric risk assessment method for risky region localization. The region riskiness is modeled by the agent and its relations to other regions. The risk probability is combined with the agent feature for a risk-embedded agent representation. The risk-embedded agent representation and the region interaction features are then fed into an agent-RNN (with a similar structure of RNN \cite{DBLP:journals/corr/ZarembaSV14}). The risky probability is measured by the Interaction of Unit (IOU) between the region locations with any ground-truth box ($>$0.4). Karim \emph{et al.} \cite{karim2022attention} improve their work \cite{karim2022dynamic} by considering the object-level riskiness score prediction, which is fulfilled by combining the hidden state of objects and their motion features together, and the riskiness is assessed by inputting the hidden state of each agent to Eq. \ref{eq:5}.

Recently, the work proposes a new dataset DRAMA \cite{DBLP:conf/wacv/MallaCDCL23} for joint risk localization and captioning in driving scenes. Different from the aforementioned works, they contribute insight into text caption prediction for important object localization. Compared with the vision input focused on in this survey,  the text description provides dense semantic guidance for future accident anticipation. By checking the keywords, such as ``crossing", the risky semantic relation between the road agents and ego-car can be assessed. Actually, for safe driving, the risk assessment of the driving scenes is very important, which can help to find the risky context in advance \cite{kataoka2020joint,DBLP:journals/tits/ChiaKGJ22,DBLP:conf/itsc/WuXCSX22}.

Most aforementioned Vision-TAA works commonly need to pre-detect the road participants in advance, and the location association models need an assumption that the number of objects in consecutive frames should be nearly consistent. As for the accident scenarios, this assumption is difficult to maintain and the severe environments make object detection not robust, which weakens the subsequent scene context and object association. In the following, we describe a detection-free Vision-TAA framework by driver attention collaboration.
 \begin{figure}[!t]
  \centering
 \includegraphics[width=\hsize]{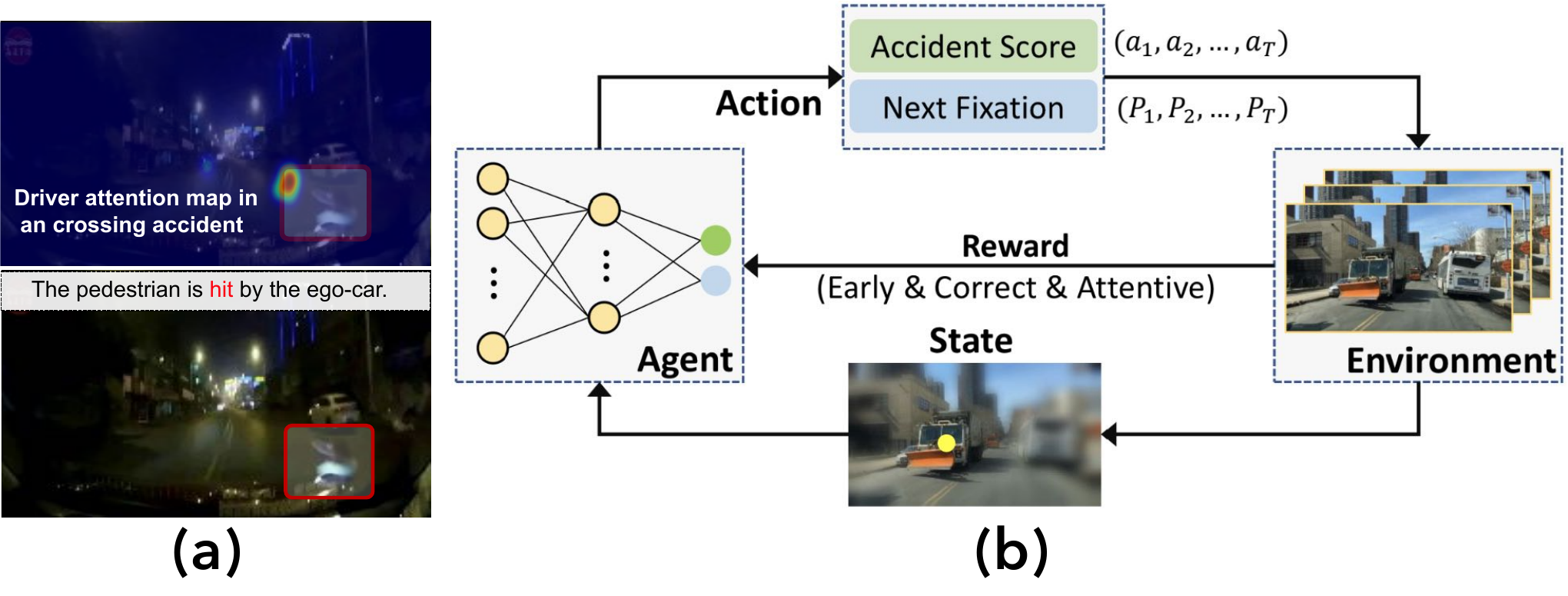}
  \caption{(a) An typical accident example with collision detection by driver attention and text assistance (from \cite{DBLP:journals/corr/abs-2212-09381}), and (b) the Markov Decision Process of accident anticipation taken from \cite{DBLP:conf/iccv/Bao0K21}.}
  \label{fig8}
\end{figure}

\subsection{Vision-TAA with Driver Attention Collaboration}
\label{taa-driver-attention}
Different from the aforementioned works, we introduce driver attention which is an active vision clue for safe driving, inspired by the mechanism of human selective attention. Commonly, driver attention also reflects the driving task indirectly because it implies where we want to go or look. 

Driver attention here can also be transferred to a kind of visual input in the form of fixation maps with the same size and frame rate as original video frames. Differently, the fixation map is treated as another output label to supervise learning spatial-temporal features in accident anticipation. This mechanism can get rid of object detection but with a promising observation for dangerous objects. The pioneering work \cite{DBLP:conf/itsc/FangYQXWL19} explores the relation between accident prediction and driver attention prediction and a positive response is obtained. Later, a work for driver attention in driving accident scenarios is proposed \cite{DBLP:journals/tits/FangYQXY22}, which shows different attention behaviors in various accident categories. 

Inspired by these works, Bao \emph{et al.} \cite{DBLP:conf/iccv/Bao0K21} propose a deep reinforced accident anticipation model with driver attention assistance in the accident window. They formulate the accident anticipation as a Markov Decision Process (MDP) fulfilled by stochastic multi-task agent optimization, \emph{i.e.}, one agent for accident prediction and another agent forecasting the driver fixations. Fig. \ref{fig8} presents the Markov Decision Process for accident anticipation assisted by driver attention.

Through the experiments in \cite{DBLP:conf/iccv/Bao0K21}, the positive role of driver attention is strongly proved with the gain of 10\% Area Under Curve (AUC) value. Besides the assistance of driver attention, Karim \emph{et al.} \cite{DBLP:journals/corr/abs-2108-00273} also validate that the attention map can boost the explainability for accident anticipation because it can focus on the dangerous regions immediately.  In our recent work, we also introduce driver attention to the accident anticipation task with a multi-task learning framework \cite{DBLP:journals/corr/abs-2212-09381}, and the positive response is further verified.

\subsection{Loss Functions}
For the accident anticipation loss, most aforementioned works adopt the Exponential Loss (ExpLoss) \cite{DBLP:conf/accv/ChanCXS16,DBLP:conf/icra/JainSKSS16} to penalize the positive accident video clips for an early accident prediction and a cross-entropy loss for the negative video clips. Hence, the ExpLoss is defined as
\begin{equation}
\mathcal{L}_a=-\sum_{t=1}^{T} a_{t}e^{-\text{max}(0,\frac{\tau}{r})}\text{log}(\hat{a}_t),
\end{equation}
where $\tau$ is the time interval of Time-to-Accident (TTA) from the current time, and $r$ is the frame rate. The ExpLoss is widely used in other accident anticipation works \cite{DBLP:conf/iccv/Bao0K21,DBLP:journals/corr/abs-2108-00273,DBLP:journals/corr/abs-2212-09381, karim2022dynamic,wang2023gsc}. Suzuki \emph{et al.} \cite{suzuki2018anticipating} extend the ExpLoss further and propose the Loss of Early Anticipation (LEA) and Adaptive Loss for Early Anticipation (AdaLEA). The difference between LEA, AdaLEA, and ExpLoss is that the exponential loss value is adaptively adjusted with the learning epoch for LEA and AdaLEA. 

For LEA, $e^{-\text{max}(0,\frac{\tau}{r})}$ in ExpLoss is changed to $e^{-\text{max}(0,\tau-\lambda(e-1))}$, where $e$ is the current learning rate, and $\lambda$ denotes a hyper-parameter for adjusting the learning rate.  LEA enforces easy-to-hard learning for traffic accident anticipation (the frames near the accident time have more accident clues than the ones far away from the accident time). AdaLEA is formulated as $e^{-\text{max}(0,\tau-r\cdot\Phi(e-1)-\eta)}$, which improves LEA further for the non-linear accident anticipation with the assistance of the mean Time-to-Accident (mTTA) \cite{DBLP:conf/accv/ChanCXS16} (to be described in the evaluation metrics in Sec. \ref{emetric}). $\Phi(.)$ represents the mTTA at each epoch, and $\eta$ is a factor of proportionality for adapting to the learning epoch. LEA and AdaLEA can fulfill an earlier penalty for the accident time.
 \begin{table}[!t]\footnotesize
	\centering
	\caption{The characteristics of representative Vision-TAA works, where the Years, Models, Inputs, and Clues are summarized.}
	\renewcommand{\arraystretch}{1.2}
	\setlength{\tabcolsep}{0.0mm}{
		\begin{tabular}{c|c|c|c|c}
			\toprule[0.8pt]
			Ref. &Years&Models&Inputs&Clues\\
			\hline\hline
			Hu \emph{et al.} \cite{hu2003traffic}&2003&3D model&Gray frames&\makecell{trajectory \\interaction}\\
			\hline
			Chan \emph{et al.} \cite{DBLP:conf/accv/ChanCXS16}&2016&DSA+RNN&RGB frames&\makecell{object\\ interaction}\\
				\hline
			Shan \emph{et al.} \cite{DBLP:journals/mms/ShanZZ17} &2014&Autoregression&vehicle position&\makecell{object \\distance}\\
				\hline
			W. Bao \emph{et al.}\cite{bao2020uncertainty}&2020&\makecell{RNN+GCN\\+BNN}&RGB frames&\makecell{object \\interaction}\\
				\hline
			H. Kim \emph{et al.}\cite{DBLP:journals/mva/KimLHS21}&2021&\makecell{Domain \\Adaptation}&\makecell{RGB frames, \\bounding boxes}&\makecell{frame clip\\ feature}\\
				\hline
			Bao \emph{et al.} \cite{DBLP:conf/iccv/Bao0K21} &2021&CNN&RGB frames& \makecell{frame-level \\hidden state}\\
				\hline
			Karim \emph{et al.} \cite{DBLP:journals/corr/abs-2108-00273}&2021&GRU&RGB frames&frame riskness\\			
			\hline
			D. Chavan \emph{et al.}\cite{DBLP:journals/corr/abs-2101-08463}&2021&Transformer&RGB frames& \makecell{object \\distance}\\
				\hline
			AV Malawade \cite{DBLP:journals/iotj/MalawadeYHMKF22}&2022&GNN+LSTM&RGB frames& \makecell{object feature\\ relationships}\\
				\hline
				Karim \emph{et al.} \cite{karim2022dynamic}&2022&DSTA+GRU&RGB frames&\makecell{frame-level \\hidden state}\\
				\hline
			Karim \emph{et al.} \cite{karim2022attention} &2023&GRU&\makecell{RGB frames,\\ optical flow}&\makecell{object \\riskiness}\\
				\hline
			Wang \emph{et al.} \cite{wang2023gsc}&2023&GCN&RGB frames&object interaction\\
				\hline
				Fang \emph{et al.} \cite{DBLP:journals/corr/abs-2212-09381}&2023&\makecell{Transformer, \\GCN, RNN}&\makecell{RGB frames,\\ driver \\attention maps}&\makecell{frame-level \\hidden state}\\
						\hline
	\end{tabular}}
	\label{tab3}
\end{table}
\subsection{Discussion}
To be clear for review, we present a summary of the attributes of the Vision-TAA methods in Table. \ref{tab3}. Based on the literature review for Vision-TAA, it is clear that the object-centric accident anticipation frameworks take the primary part, while the driver attention-assisted Vision-TAA provides a kind of interpretation for why the objects are focused under critical situations. In addition, besides borrowing the trajectory prediction models, the research pipelines for object-centric Vision-TAA are mainly inspired by the work \cite{DBLP:conf/accv/ChanCXS16}, which formulates the Vision-TAA as a classification problem. Compared with the model diversity in the Vision-TAD task, the formulation of Vision-TAA is more consistent, and the ExpLoss is commonly adopted with an early awareness of accidents for the positive samples. In the meantime, the accident window labels are utilized for the supervised Vision-TAA. This type of formulation causes a \emph{background confounding} issue \cite{li2022invariant} that the accident's causal part may not be learned well for accident anticipation and easily be influenced by the normal background frames. 
 \begin{table*}[!t]\footnotesize
  \centering
  \caption{The available datasets for Vision-TAD with the attributes of the number of sequences (Seq. num), citations (Cites.), annotations, accident categories (Categ.), synthetic (S) or real (R), observation views (dashcam, surveillance, and BEV), and dataset links.}
\setlength{\tabcolsep}{0.2mm}{
\begin{tabular}{c|c|c|c|c|c|c|c|l}
\toprule[0.8pt]
Datasets&Years&Cites.&Seq. num&Annotations&Categ.&S/R& Obs. View&URL\\
\hline\hline
\textcolor{magenta}{A3D} \cite{DBLP:conf/iros/YaoXWCA19}&2019 &113&1500&T&-& R& Dashcam&\url{https://github.com/MoonBlvd/tad-IROS2019}\\
\textcolor{magenta}{DVAD} \cite{DBLP:conf/icccnt/IjjinaS19}&2019 &9&60&T, Text&-& R&Dashcam&not available\\
\textcolor{magenta}{Drive-Anomaly106} \cite{DBLP:journals/sensors/ZhuFXX19}&2019 &4&106&T, S&-& R&Dashcam&\url{https://github.com/ZHU912010/Driving-Anomaly-Detection}\\
\textcolor{magenta}{RetroTrucks} \cite{DBLP:conf/ivs/HareshKZT20} &2020&21&474&T, S&\checkmark&R&Dashcam&Long link*\\\
\textcolor{magenta}{TNAD} \cite{DBLP:journals/tsas/HuangHRR20}&2020&76&106&T, S& \checkmark& R&BEV&\url{https://ankitshah009.github.io/accident_forecasting_traffic_camera}\\
\textcolor{magenta}{ADV} \cite{le2020attention}&2020&18&100&T, S&-& R&Dashcam&\url{https://sites.google.com/view/ltnghia/research/accident-detection}\\
\textcolor{magenta}{TAD-1} \cite{DBLP:journals/tip/LvZCXLY21}&2021 &55&500&T, S& \checkmark& R&Dashcam, Sur.&\url{ https://github.com/ktr-hubrt/WSAL}\\
\textcolor{magenta}{TaskFix} \cite{juan2021investigating}&2021 &0&1436F*&T, Att.& -& R&Sur.&\url{https://bit.ly/TaskFixDataset}\\
\textcolor{magenta}{USDC} \cite{yawovi2022wrong}&2022 &0&122&T, S&-& R& Dashcam&\url{https://public.roboflow.com/object-detection/self-driving-car}\\
\textcolor{magenta}{TrafficS} \cite{ghahremannezhad2022real}&2022 &4&29&T, S&\checkmark& R& Sur.&\url{http://github.com/hadi-ghnd/AccidentDetection}\\
\textcolor{magenta}{DADA-2000} \cite{DBLP:journals/tits/FangYQXY22}&2022&50  &2000&T, Text, Att.& \checkmark& R&Dashcam&\url{https://github.com/JWFangit/LOTVS-DADA}\\
\textcolor{magenta}{DoTA} \cite{yao2022dota}&2022 &13&4,677&T, S& \checkmark& R& Dashcam&\url{https://github.com/MoonBlvd/Detection-of-Traffic-Anomaly}\\
\textcolor{magenta}{TAD-2} \cite{xu2022tad}&2022&2 &333&T, S& \checkmark& R&Sur.&\url{https://github.com/yajunbaby/TAD-benchmark}\\
\textcolor{magenta}{DADA-seg} \cite{DBLP:journals/tits/ZhangYS22}&2022&11&313&T, SEG&-& R&Dashcam&\url{https://github.com/
jamycheung/ISSAFE}\\
\textcolor{magenta}{MP-RAD} \cite{DBLP:journals/tits/VijayDCNK23}&2023 &0&2000&T&-& S&Sur.&\url{https://github.com/draxler1/MP-RAD-Dataset-ITS-}\\
\textcolor{magenta}{DRAMA} \cite{DBLP:conf/wacv/MallaCDCL23}&2023 &2&17,785&Risk, S, Text& \checkmark& R& Dashcam&\url{https://usa.honda-ri.com/drama}\\
\textcolor{magenta}{CTAD} \cite{luoICASSP2023}&2023 &0&1100&T&& S& Sur.&\url{https://github.com/hankluo2/UrbanTrafficAccidentDetection}\\
  \hline
  \end{tabular}}
\begin{tablenotes}
\item \scriptsize{PSJ: Philippine Journal of Science; ICIST: IEEE International Conference on Imaging Systems and Techniques. 1436F: 1436 frames. Long link: \url{https://drive.google.com/drive/folders/1VxFG1jHBiep4R3i_MmvMfKWH11AEFFhu}. The citations (Cites.) were reported by Google Scholar on July 25, 2023.}
\end{tablenotes}
  \label{tab4}
  \end{table*}

 \begin{table*}[!t]\footnotesize
  \centering
  \caption{The available datasets for Vision-TAA with the attributes of the number of sequences (Seq. num), citations (Cites.), annotations, accident categories (Categ.), synthetic (S) or real (R), observation views (dashcam, surveillance, and BEV), and dataset links.}
\setlength{\tabcolsep}{0.5mm}{
\begin{tabular}{c|c|c|c|c|c|c|c|l}
\toprule[0.8pt]
Datasets&Years&Cites.&Seq. num&Annotations&Categ.&S/R& Obs. View& URL\\
\hline\hline
\textcolor{magenta}{DAD} \cite{DBLP:conf/accv/ChanCXS16}&2016&196&3000&T&-& R&Dashcam&\url{http://aliensunmin.github.io/project/dashcam/}\\\
\textcolor{magenta}{Epic Fail} \cite{DBLP:conf/cvpr/ZengCCNS17} &2017&54&1,750&T&-& R&Dashcam&\url{http://aliensunmin.github.io/project/video-Forecasting/}\\\
\textcolor{magenta}{NIDB} \cite{DBLP:conf/icra/KataokaSOMS18}&2018&31&6,200&T&-& R&Dashcam& not available\\
\textcolor{magenta}{CADP} \cite{shah2018accident}&2018&79&1416&T, S& \checkmark& R&Sur.&\url{https://ankitshah009.github.io/accident_forecasting_traffic_camera}\\
\textcolor{magenta}{VIENA$^2$} \cite{aliakbarian2019viena}&2019&29&15,000&T& \checkmark& S&Dashcam&\url{https://sites.google.com/view/viena2-project/home}\\
\textcolor{magenta}{GTACrash} \cite{DBLP:conf/aaai/KimLHS19}&2019 &25&7,720&T, S&-& S&Dashcam&\url{https://github.com/gnsrla12/predicting-vehicle}\\
\textcolor{magenta}{CTA} \cite{DBLP:conf/eccv/YouH20}&2020 &25&1,935&T& \checkmark& R&Dashcam&\url{https://github.com/tackgeun/CausalityInTrafficAccident}\\
\textcolor{magenta}{CCD} \cite{bao2020uncertainty}&2021&55 &1,500&T, Text&-& R&Dashcam&\url{https://github.com/Cogito2012/CarCrashDataset}\\
\textcolor{magenta}{YouTubeCrash} \cite{DBLP:journals/mva/KimLHS21}&2021&3 &122&T, S&-& R& Dashcam&\url{-collisions-using-data-collected-from-video-games}\\
\textcolor{magenta}{SUTD-TrafficQA} \cite{DBLP:conf/cvpr/XuHL21}&2021 &39&10,080&T, Text& \checkmark& R&Sur.&\url{https://github.com/SUTDCV/SUTD-TrafficQA}\\
\textcolor{magenta}{TRA} \cite{DBLP:journals/tits/LiuLCLX22}&2022 &4&2000&Risk&\checkmark& R& Dashcam&\url{hhttps://github.com/liuchunsense/TRA}\\
\textcolor{magenta}{DADA-2000} \cite{DBLP:journals/tits/FangYQXY22} &2022&50&2000&T, Text, Att.& \checkmark& R&Dashcam&\url{https://github.com/JWFangit/LOTVS-DADA}\\
\textcolor{magenta}{CAP} \cite{DBLP:journals/corr/abs-2212-09381}&2022 &5 &11,727&T, Text& \checkmark& R&Dashcam&\url{https://github.com/JWFanggit/LOTVS-CAP}\\
\textcolor{magenta}{ROL} \cite{karim2022attention}&2023 &1 &1000&T, S&-& R& Dashcam& not available\\
\textcolor{magenta}{DeepAccident} \cite{DBLP:journals/corr/abs-2304-01168}&2023&0 &57K frames&T, S& \checkmark& S&Dashcam&\url{https://github.com/JWFanggit/LOTVS-CAP}\\
  \hline
  \end{tabular}}
  \begin{tablenotes}
\item \scriptsize{The citations (Cites.) were reported by Google Scholar on July 25, 2023.}
\end{tablenotes}
  \label{tab5}
  \end{table*}
  
\section{Available Datasets}

\begin{figure}[!t]
  \centering
 \includegraphics[width=\hsize]{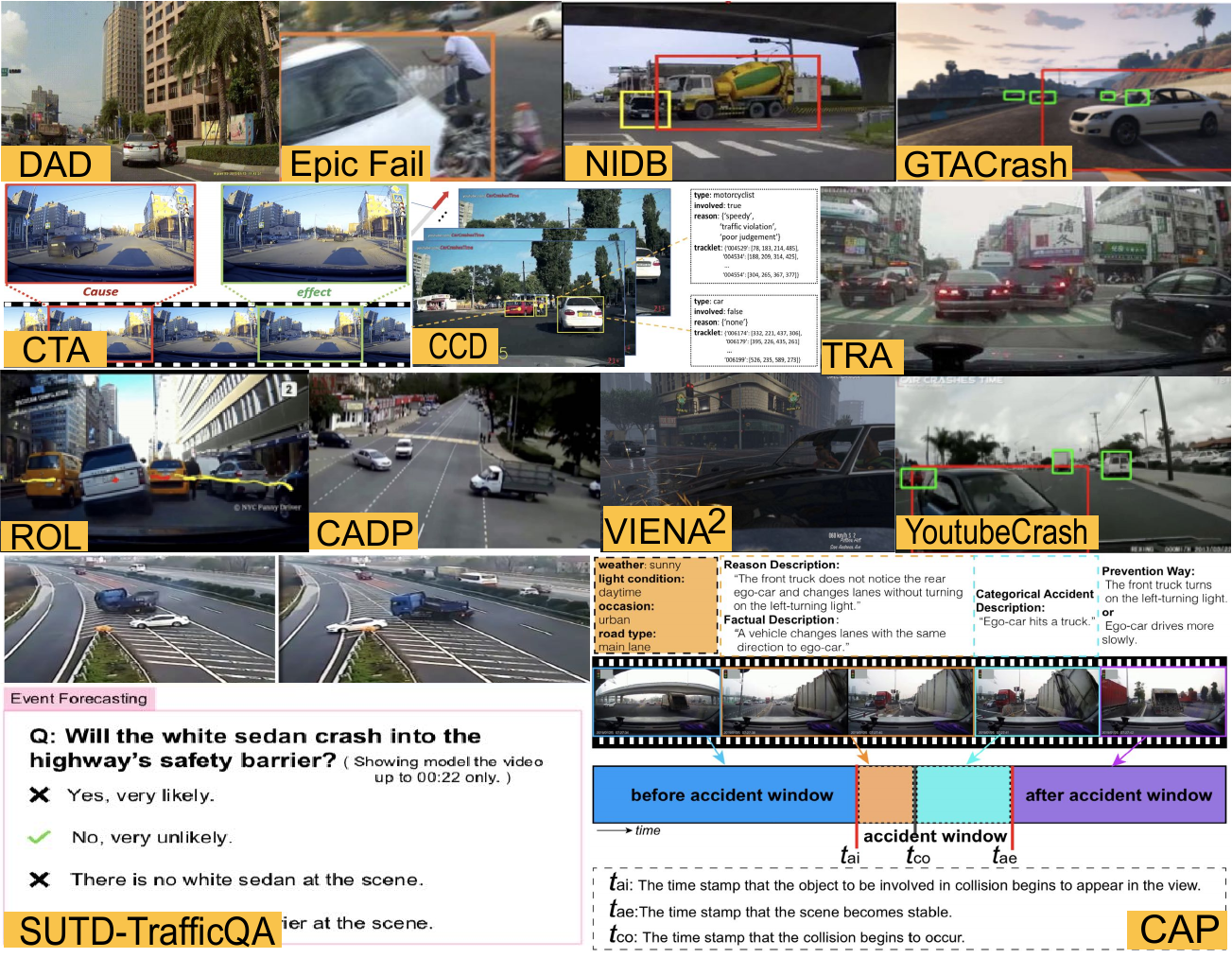}
  \caption{The samples in traffic accident datasets for Vision-TAA task.}
  \label{fig9}
\end{figure}

From the aforementioned works, many researchers are contributing excellent and hard efforts for Vision-TAD and Vision-TAA tasks, and many datasets have been developed in recent years. Table. \ref{tab4} and Table. \ref{tab5} presents the related datasets for Vision-TAD and Vision-TAA tasks, respectively. The downloading links for these datasets are all provided for quick utilization and review. To be clear, we also exhibit the snapshots of the Vision-TAA datasets in Fig. \ref{fig9}.
  
\subsection{Description of Some Representative Datasets}
Some representative datasets are depicted as follows. 

\textbf{DoTA} \cite{yao2022dota} is constructed by the Robotics Institute at the University of Michigan, which contains 4,677 dashcam accident videos with 731,932 frames (10fps). DoTA is extended from the A3D \cite{DBLP:conf/iros/YaoXWCA19} in the same team. It annotates the temporal accident window, object location tracklets, and accident categories. The categories contain nine accident categories from the movement direction of vehicles, pedestrians, and other roadblocks. The primary object category is the car (73.2\%), and ego-car-involved accidents take about 51.9\%.  

\textbf{CADP} \cite{shah2018accident} contains 1416 surveillance video segments (accidents) collected from YouTube by Language Technologies Institute at Carnegie Mellon University, where 205 video segments own the full spatial-temporal annotation. The average video length in CADP is about 366 frames, which implies an average Time-to-Accident (TTA) of 3.69 seconds. Similarly, cars take the primary object type. CADP can be used for Vision-TAD and Vision-TAA tasks.

\textbf{DAD} \cite{DBLP:conf/accv/ChanCXS16} is the first widely used dataset for dashcam accident prediction collected by the VSLab at the National Tsing Hua University, Taiwan. It mainly concentrates on motorbike-involved dashcam accidents, where the 678 videos in DAD are gathered from Taipei, Hsinchu, Taichung, Kaohsiung, Yilan, and Tainan cities. These 678 videos are sampled into 1750 clips (pos: 620 and neg: 1130), containing 100 frames for each clip. About ten accident frames are trimmed at the end of each clip for the positive accident clips.

\textbf{CCD} \cite{bao2020uncertainty} is developed by Rochester Institute of Technology, USA. It contains 1500 dashcam accident videos and each video annotates the accident window, object locations, accident reasons, and some other weather and light conditions. Differently, all accidents involve car collisions. The object locations are obtained by Cascade R-CNN \cite{cai2018cascade}. Each video in CCD contains 50 frames for covering 5 seconds of time.

\textbf{DADA-2000} \cite{DBLP:journals/tits/FangYQXY22} is collected by Chang'an University and Xi'an Jiaotong University, China. It contains 2000 accident videos with 658, 476 frames, and the average video length is about 230 frames (30fps). Besides the temporal annotation, DADA-2000 gathers the driver attention data for each video frame on 54 kinds of accidents. It is still the largest dataset for driver attention prediction in accident scenarios and is used for attention-assisted accident anticipation in DRIVE \cite{DBLP:conf/iccv/Bao0K21}.

\textbf{SUTD-TrafficQA} \cite{DBLP:conf/accv/ChanCXS16} focuses on the traffic accident video Question Answering (QA) for the text description answer for basic understanding, event forecasting, reverse reasoning, counterfactual inference, and introspection. Different from the aforementioned datasets, SUTD-TrafficQA is constructed by the Singapore University of Technology and Design, which prefers a high-level understanding of the accident and formulates the accident prediction as a question-answering problem.

\textbf{GTACrash} \cite{DBLP:conf/aaai/KimLHS19} is a synthetic dataset generated by GTV-5, which is contributed by Krafton Inc. Seongnam, Korea Advanced Institute of Science and Technology, and University of Wisconsin-Madison, USA. GTACrash also focuses on dashcam car collisions, owning 7,720 accident videos (10fps) and 3661 non-accident videos. With the help of GTV-5, accurate bounding boxes and the tracklets of objects are obtained conveniently. GTACrash can be used to improve Vision-TAA in real situations with domain adaptation approaches.

\subsection{Critical Review of Datasets}
With the investigation, the attributes of these datasets can be reviewed from observation views, annotation diversity, and the necessity of Vision-TAD and Vision-TAA.

\subsubsection{Observation Views}
From these tables, we can see that the datasets can be divided into three different views: BEV view, dashcam view, and surveillance view. Observation view is a direct influencing factor for Vision-TAD and Vision-TAA. Road participants exhibit various evolution processes because of the difference in scale change, occlusion frequency, and camera motion variation. The specific attributes of different views are as follows. In the dashcam view, the object scale varies greatly, which needs the object detector (pre-step for accident detection or prediction) that can adapt to the severe scale change and occlusion better than the BEV and surveillance views. BEV view has minimal scale change and occlusion issues.  Surveillance view is commonly stationary~\cite{zheng2020vehiclenet}, while BEV view sometimes is captured by unmanned drones that dynamically move~\cite{zheng2020university}. Dashcam view mounted on the vehicles has severe camera motion, which hardens the challenges for Vision-TAD and Vision-TAA.
	
In addition, different observation views have different degrees of physical distortion, which is caused by the view plane coordinate system. For certain objects, the observation direction of the BEV view may be orthogonal to the movement directions of the road participants, and the trajectories in the BEV view can be measured with little physical distortion.

\subsubsection{Annotation Diversity}
In the early stage, the datasets for Vision-TAD commonly annotate the spatial object locations and temporal windows for accidents, and provide spatial-temporal accident localization. On the contrary, Vision-TAA concentrates on the earliness of accident anticipation. Therefore, many datasets in early Vision-TAA only annotate the temporal accident window.
With the development of deep learning models, small-scale datasets are not enough to fulfill a well-trained model. Therefore, the large-scale datasets with spatial location annotation for Vision-TAA emerge frequently util 2022. However, because of the problem setting of Vision-TAA, temporal annotation is still dominant. 

Recently, some Vision-TAA datasets provide the text description with text-vision multimodality development. Of course, this work focuses on a vision-based accident understanding. In fact, we have a prediction that multimodality datasets \cite{DBLP:journals/eswa/ChoiKKL21} will appear more and more because of the progress of Artificial Intelligence Generated Content (AIGC)-based techniques. DeepAccident \cite{DBLP:journals/corr/abs-2304-01168} is the only accident dataset in V2X scenarios, while differently, it focuses on an object-level accident prediction with a threshold of 2.5 meters of objects for collision warning. In self-driving systems, the current configuration takes the 3D LiDAR as the popular sensor. However, there is no large-scale accident dataset that contains the 3D LiDAR data mainly because it is rather difficult to collect enough accident data by the 3D LiDAR sensor in practical scenarios. For example, in January 2023, Waymo's self-driving cars reached 1 million miles, and only two crashes were involved \cite{waymo2023}. Maybe, in the future, the virtual simulation may be the main choice for accident understanding collaborating 3D LiDAR data.

\subsubsection{Necessity of Vision-TAD and Vision-TAA}
From Table. \ref{tab5}, dashcam view takes the main part in Vision-TAA datasets because of the high possibility of collision avoidance by controlling vehicles.  If we want to add the accident avoidance ability to BEV or surveillance views, we need to consider the video anomaly detection \cite{DBLP:journals/tip/LvZCXLY21,DBLP:conf/ijcai/0030ZLW0LDL21} for finding suspicious behaviors before the accident or consider a wide-range road network for road-level accident prediction \cite{DBLP:journals/tits/LiyanageZC22,DBLP:conf/itsc/YangWZZW20}. Certainly, many works in dashcam view also conduct the Vision-TAD \cite{DBLP:conf/ivs/HareshKZT20,DBLP:conf/iros/YaoXWCA19,yao2022dota}, mainly for post-accident analysis or ego-car uninvolved accident avoidance. In addition, from the citations of the datasets, we can see that vision-based traffic accident detection and anticipation are still understudied, and Vision-TAA is \textbf{more attractive} than Vision-TAD.

\subsection{Evaluation Metrics}
\label{emetric}

Through the progress, Vision-TAD and Vision-TAA have formed common evaluation metrics. 

\subsubsection{Metrics for Vision-TAD}  Most metrics for Vision-TAD follow the ones in video anomaly works \cite{DBLP:conf/cvpr/SultaniCS18,DBLP:journals/corr/abs-2211-15098}, but with a different object location or temporal window annotation. If we want to evaluate the spatial and temporal accident localization performance, the frame-level Receiver Operating Characteristic (\textbf{ROC}) curve and Area under the ROC curve (\textbf{AUC}) are commonly used, which evaluate the sensitivity of methods for localizing the temporal accident frames. 

Frame-level ROC curves contain two axes, \emph{i.e.}, True Positive Rate (TPR) and False Positive Rate (FPR), which computes the ratio between truly detected accident frames, and falsely detected ones with the ground truth, respectively. AUC calculates the area under the ROC curve, and larger AUC prefers a better performance. Besides the frame-level evaluation, some works focus on the spatial accident region localization further, which results in a spatial-temporal detection evaluation. For example, DoTA \cite{yao2022dota} presents a Spatial-Temporal Area Under Curve (\textbf{STAUC}), which computes the True Accident Regions Rate (TARR) and the Spatial-Temporal TPR (STTPR) by
\begin{equation}
\text{TARR} = \frac{\sum_{p\epsilon m_t}\Delta I(p)}{\sum_{p\epsilon M}\Delta I(p)},
\end{equation}
and
\begin{equation}
 \text{STTPR} = \frac{\sum_{t\epsilon TP}\text{TARR}_t}{|TP|},
\end{equation}
where $\Delta I(p)$ denotes the accident score at pixel $p$, $M$ and $m_t$ denote the whole frame and the annotated accident region at time $t$, respectively. $TP$ denotes all ground truth accident frames. Based on the computation of STTPR and FPR, the STAUC can be obtained, which considers the spatial accident score arrangement issue over video sequence. The best AUC and STAUC values on the DoTA dataset are 73.0\% and 48.5\% currently \cite{yao2022dota}. It implies a large space to be improved.

In addition to the ROC family, \textbf{Precision}, \textbf{Recall}, and \textbf{F1}-measure are also the hot choices for the Vision-TAD task. Similar to object detection works, Precision and Recall evaluate the truly detected accident frames or local regions with threshold partitioning. After the separation by a certain threshold, the binary results can be evaluated with the ground truth. Precision computes the overlapping ratio of detected positive results with all positive ones, and Recall calculates the covering ratio between the detected results with all the positive and negative items (frames or pixels).

Certainly, some works advocate an Intersection of Unit (IOU) evaluation for temporal accident frames \cite{DBLP:conf/eccv/YouH20} or spatial accident regions \cite{yao2022dota,DBLP:journals/corr/abs-2304-01168}. IOU evaluates the overlapping rate between the detected positive samples with the ground truth, which belongs to the metric system of Precision and Recall with a binary partitioning. 

\subsubsection{Metrics for Vision-TAA} 
The metrics for Vision-TAA contain three common ones: \emph{i.e.}, Average Precision (AP) \cite{bao2020uncertainty}, and Time-to-Accident (TTA) \cite{DBLP:conf/iccv/Bao0K21} with the variants of TTA$_{a_c}$ and mTTA. 
\textbf{TTA}$_{a_c}$ evaluates the earliness of positive accident anticipation by setting the accident score threshold in each video as $a_c$ (set as 0.5 generally). mTTA computes the mean TTA by changing $a_c$ from 0 to 1. 
\textbf{AP} is computed by the precision values. As aforementioned, precision and recall need a threshold to binarize the results into positive and negative results. AP evaluates the average precision with different thresholds of precision rate. 
\textbf{AUC} is also a metric for evaluating video-level accident classification performance, where the positive or negative classification is determined by whether there is a frame with the accident occurrence probability larger than $a_c$. These metrics prefer larger values for better performance. 
AP and TTA metrics are focused on the evaluation of accident prediction. However, based on our investigation in \cite{DBLP:journals/corr/abs-2212-09381}, we find that the AP metric is easily influenced by the number of negative samples and more negative samples will degrade the AP values. Besides, mTTA shows weak performance evaluation ability, where different configurations may generate similar mTTA values. TTA$_{a_c}$ and AUC can provide a fair evaluation for different works.

The special metrics of the TTA ones (\emph{i.e.}, \text{TTA$_{a_c}$} and mTTA) are computed by
\begin{equation}
 \text{TTA}_{a_c}= \max{\{\tau|a_t\geq{a_c},1\le t \le t+\tau}\},
\end{equation}
and
\begin{equation}
 \text{mTTA} = E(\text{TTA}),
\end{equation}
where $a_t$, $a_c$ denote accident prediction scores and accident score thresholds respectively, and $E(.)$ computes the Expectation of  \text{TTA}$_{a_c}$ with different $a_c\in[0,1]$. 

Through an investigation by ourselves and the work \cite{bao2020uncertainty}, TTA metrics determine the time-to-accident interval by the first timestamp which obtains an accident score larger than $a_c$. It cannot evaluate the anticipation stability of the anticipated accident scores. For a safe-driving system, it is indispensable to avoid frequent false alarms.  Therefore, the work of \cite{bao2020uncertainty} estimates the mean Algebraic Uncertainty (mAU) and mean Epistemic Uncertainty (mEU) for the anticipated accident score at time $t$ by the formulation of
\begin{equation}\small
{\bf{U}}_t=\underbrace{\frac{1}{K}\sum_{k=1}^{K}[\text{diag}(\hat{a}_k)-\hat{a}_k\hat{a}_k^T]}_{\text{mAU}}+\underbrace{\frac{1}{K}\sum_{k=1}^{K}(\hat{a}_k-\bar{a})(\hat{a}_k-\bar{a})^T}_{\text{mEU}},
\end{equation}
where ${\bf{U}}_t$ is the anticipation uncertainty at time $t$, $\hat{a}_k$ is the video accident score at time $t$ for the $k$-th forward pass of anticipation, and $\bar{a}=\frac{1}{K}\sum_{k=1}^{K}\hat{a}_k$ over $K$ times of prediction, where $\hat{a}_k=(\hat{a}^{n}_k,\hat{a}^{p}_k)$ contains the accident score for the negative sample and positive sample at time $t$ in $k$-th forward prediction. 

In addition, current Vision-TAA generally tests the performance on the video sequence with a fixed length (\emph{e.g.}, 150 frames). In practical use, we may implement the Vision-TAA model in online testing. Therefore, the metrics for Online Vision-TAA evaluation are needed in future.

\section{Future Insights}
\label{trend}
We next discuss the research insights and trends for vision-based traffic accident detection and anticipation.

\textbf{1) Toward Explainable Vision-TAD and Vision-TAA:}  As a commonplace, explainability is one of the core issues for Vision-TAD and Vision-TAA tasks because of the safe-critical property. For surveillance systems, every alarm of an accident may cause high-cost measures for accident cleaning. Therefore, many surveillance recording systems still arrange many humans in the monitoring hall currently. Therefore, Vision-TAD needs to know why a false alarm is caused by the data or model aspects, so as to re-design or re-organize the model architecture. As for safe-driving systems, Vision-TAD can be commonly used to localize the accident video clips for accident responsibility investigation. Vision-TAA mainly works for safe-driving systems to avoid accidents. For this purpose, the prediction uncertainty makes the Vision-TAA model explainable for reducing false alarms because every accident warning may change the decision of driving systems.

\textbf{2) Scalable and Causal Vision-TAD and Vision-TAA:} Currently collected accident benchmarks are actually the closed accident world. The diversity and accident patterns are fixed in each benchmark. In the open-traffic world, every road participant faces new and unseen traffic situations every time. This is more universal to the Vision-TAD and Vision-TAA models compared with normal situations. Therefore, the scalability of Vision-TAD and Vision-TAA is one of the core issues for future research. For unseen accident scenarios, domain adaptation~\cite{DBLP:journals/pami/ZhouLQXL23}, or scenario generation, such as stable diffusion \cite{rombach2022high}, may be promising by borrowing the knowledge of large vision models.  In addition, accidents in videos commonly occupy a very short window, which implies a confounding issue for the causal grounding of Vision-TAD and Vision-TAA. Therefore, some invariant video grounding \cite{li2022invariant} or counterfactual thinking approaches are helpful. 

\textbf{3) Label Efficient Learning for Vision-TAD:} Because of the long-tailed property, it is impossible to collect the whole traffic accident world, and the generalization of Vision-TAD on unseen accident data will still be challenging. Commonly, we can only provide the annotation for very few accident data. Therefore, if we want to fully leverage the few-shot annotation to the unseen or other unlabeled accident videos, label efficient learning, such as active learning \cite{assran2022masked, ji2023are}, may be a promising approach. By learning the invariant representation among different accident videos, the expected adaptation ability over various situations is preferred.

\textbf{4) Multi-modal fusion for Vision-TAD and Vision-TAA:} The common paradigms concentrate on the consistency or correlation of the object tracklets and video frames. However, it is common sense that accidents in harsh environmental conditions increase the difficulty of detecting or tracking the target object, and the feasibility of these mainstream methods is limited. In safe-driving systems, LiDAR information \cite{DBLP:journals/corr/abs-2202-02703} and sound \cite{DBLP:journals/eswa/ChoiKKL21} are all helpful for perceiving the distances. Therefore, RGB-X (\emph{e.g.}, \emph{text, sound, point cloud, etc}.) fusion can be explored in the future from the new benchmarks and promising models.

\textbf{5) Borrowing Real-to-Sim-to-Real (RSR) Framework:} Because of the safe-critical and long-tailed property, few research works begin to borrow the help of synthetic or simulated data. Inspired by this insight, we think a Real-to-Sim-to-Real (RSR) framework may be more appropriate. Due to the domain distribution gap between synthetic and real data, directly generating the data only by human imagination is not adequate. This implies hard efforts to reduce the domain gap. With the increase of accident benchmarks collected in the real world, if we can transfer the knowledge of real accidents to synthetic data (R2S), the realistic distribution may be transferred to the virtual accident world. In addition, with the help of controllable annotation and the diversity of virtual accidents, we can further transfer the diversity and annotation to the real accident world (S2R). By this circle-type RSR framework, the model scalability can be largely boosted because it will cover the accident types as much as possible.

\textbf{6) Vision-TAD and Vision-TAA in V2X Scenarios:} Current Vision-TAD and Vision-TAA work in dashcam view concentrate on the single observation view, where the observation range and distance are limited. Based on the fast development of connected vehicles, V2X scenarios \cite{xu2022v2x,DBLP:journals/corr/abs-2304-01168} certainly will be the key situation. This trend implies many valuable research insights, such as the shared accident scene reconstruction (\emph{e.g.,} recovering BEV view of accidents), consistent accident projection among vehicles, and 3D accident scenarios construction. In addition, the accident detection results may be shared with different vehicles. Therefore, a lot of research points can be derived in V2X scenarios.

 \begin{figure}[!t]
  \centering
 \includegraphics[width=0.9\hsize]{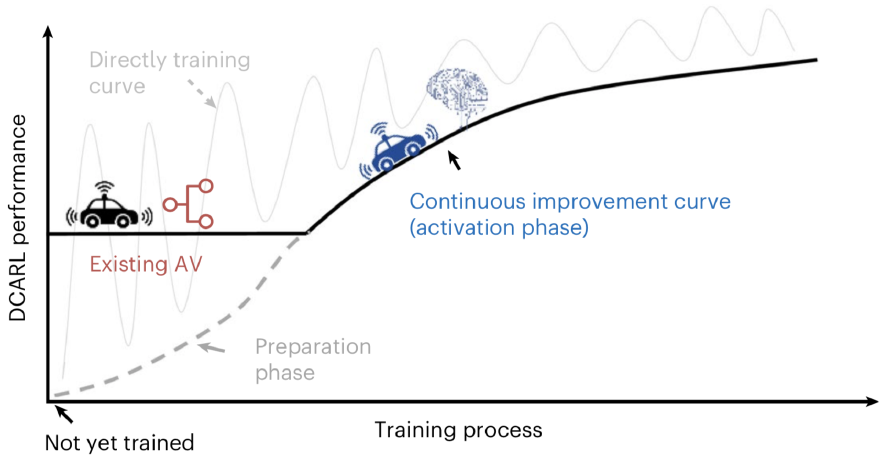}
  \caption{The DCARL framework in \cite{cao2023continuous}.}
  \label{fig10}
\end{figure}

\textbf{7) Continual Reinforcement Vision-TAA:} The ultimate goal of Vision-TAA is to avoid road collisions. As an active problem, Vision-TAA involves uncertainty estimation, risk assessment, decision, and planning behaviors. Facing the dynamic, complex driving scenes, we need to make the Vision-TAA model fulfill a self-reinforcement with the computing of large-scale sequential video data, so as to adapt the active accident prediction. Recently, Zhong \emph{et al.} \cite{cao2023continuous} developed a Dynamic Confidence-Aware Reinforcement Learning (DCARL) technology for the continuous improvement of safe driving, as shown in Fig. \ref{fig10}. Can this formulation be feasible for the Vision-TAA task? The positive answer will be promising.

\section{Conclusion}
With the full portrait description, we investigate vision-based traffic accident detection and anticipation from the aspects of challenges, data factors, main research pipelines, and available benchmarks. In addition, some issue discussions in each part are summarized. The main problem is that most current methods for Vision-TAD still follow the framework of video anomaly detection, and the distribution confusion between general traffic anomaly and accidents is not clearly addressed. In addition, most Vision-TAA works formulate a supervised classification problem, which lacks the online testing ability for sequentially upcoming traffic video frames. Based on the survey, we find that there is still vast work to be done for surveillance safety and safe driving. From this survey, we hope Vision-TAD and Vision-TAA problems can bring springing progress from effective models, new benchmarks, insights, and practical applications.

{\small
\bibliographystyle{IEEEtran}
\bibliography{ref}
}

\end{document}